%% file: main.tex
\documentclass[sigconf]{acmart}
\pdfoutput=1
\AtBeginDocument{%
  \providecommand\BibTeX{{%
    \normalfont B\kern-0.5em{\scshape i\kern-0.25em b}\kern-0.8em\TeX}}}

\usepackage{times}  % DO NOT CHANGE THIS
\usepackage{helvet}  % DO NOT CHANGE THIS
\usepackage{courier}  % DO NOT CHANGE THIS
\usepackage{graphicx} % DO NOT CHANGE THIS
\usepackage[noend]{algpseudocode}
\usepackage{algorithmicx}
\usepackage{natbib}  % DO NOT CHANGE THIS AND DO NOT ADD ANY OPTIONS TO IT
\usepackage{caption} % DO NOT CHANGE THIS AND DO NOT ADD ANY OPTIONS TO IT
\usepackage{slashbox}

\usepackage[ruled,vlined, linesnumbered]{algorithm2e}
\usepackage{subcaption}
\usepackage{amsmath}
\usepackage{amsfonts}
\usepackage{slashbox}

\SetKwInOut{KwIn}{Input}
\SetKwInOut{KwOut}{Output}
\newcommand{\argmin}{\mathop{\rm argmin}}

\newtheorem{theorem}{Theorem}

\begin{document}

\title{Wasserstein Adversarial Examples on Univariant Time Series Data}




\author{Wenjie Wang}
\affiliation{%
  \institution{Emory University}
  \city{Atlanta}
  \country{USA}}
\email{wang.wenjie@emory.edu}

\author{Li Xiong}
\affiliation{%
  \institution{Emory University}
  \city{Atlanta}
  \country{USA}}
\email{lxiong@emory.edu}

\author{Jian Lou}
\affiliation{%
  \institution{Zhejiang University}
  \city{Hangzhou}
  \country{China}}
\email{jian.lou@zju.edu.cn}

\begin{abstract}

Adversarial examples are crafted by adding indistinguishable perturbations to normal examples in order to fool a well-trained deep learning model to misclassify. In the context of computer vision, this notion of indistinguishability is typically bounded by $L_{\infty}$ or other norms.  However, these norms are not appropriate for measuring indistinguishiability for time series data. In this work, we propose adversarial examples in the Wasserstein space for time series data for the first time and utilize Wasserstein distance to bound the perturbation between normal examples and adversarial examples. We introduce Wasserstein projected gradient descent (WPGD), an adversarial attack method for perturbing univariant time series data. We leverage the closed-form solution of Wasserstein distance in the 1D space to calculate the projection step of WPGD efficiently with the gradient descent method. We further propose a two-step projection so that the search of adversarial examples in the Wasserstein space is guided and constrained by Euclidean norms to yield more effective and imperceptible perturbations. We empirically evaluate the proposed attack on several time series datasets in the healthcare domain. Extensive results demonstrate that the Wasserstein attack is powerful and can successfully attack most of the target classifiers with a high attack success rate. To better study the nature of Wasserstein adversarial example, we evaluate a strong defense mechanism named Wasserstein smoothing for potential certified robustness defense. Although the defense can achieve some accuracy gain, it still has limitations in many cases and leaves space for developing a stronger certified robustness method to Wasserstein adversarial examples on univariant time series data.
\end{abstract}
\iffalse


\fi

\maketitle

\input{intro}

\input{related}

\input{method}

\input{results}

\input{conclusion}
\bibliographystyle{ACM-Reference-Format}
\small
\bibliography{main}

\appendix

\end{document}

%% file: intro.tex
\vspace{-0.2cm}
\section{Introduction}

Deep learning has been widely applied to real-world time series classification tasks including healthcare, power consumption monitoring, as well as social safety and security \cite{TSC-review, 7552276, Ma2018HealthATMAD}. Univariant time series modeling is a common class of applications that deals with one observed variable over time \cite{karim2019multivariate}.  The data are often collected from a sensor, such as Electrocardiography (ECG) to measure patient's heart rhythms \cite{castells2007principal}, and can be used for diagnosis and monitoring. %Many such applications are  security-concerned. 
Deep learning systems have  achieved state-of-the-art performance in most of these tasks.
\vspace{-0.1cm}
\begin{figure}[ht]
  \centering
  \setlength{\abovecaptionskip}{0.cm}
  \includegraphics[width=0.5\textwidth]{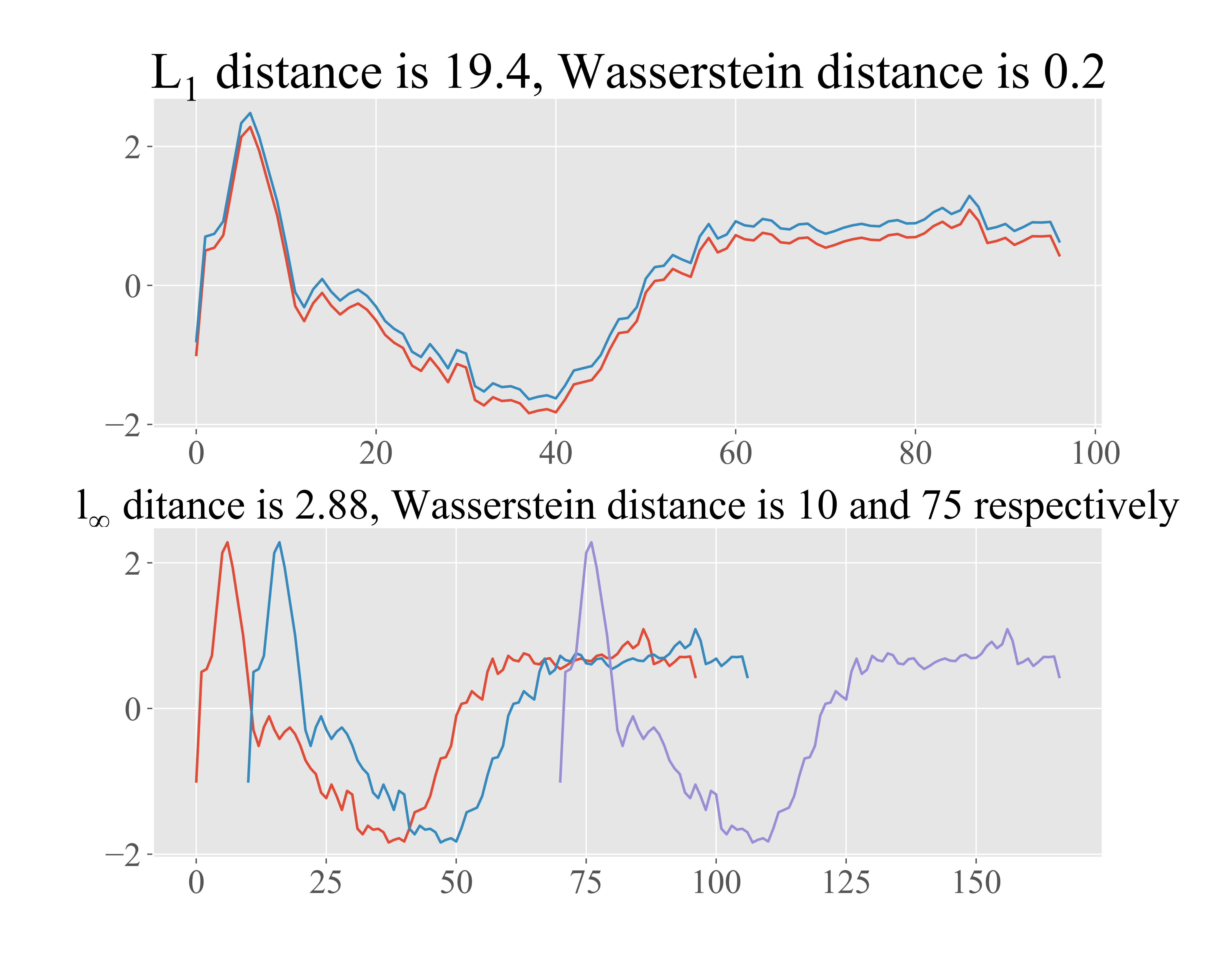}
  \caption{An example to illustrate the differences between Wasserstein distance and Euclidean distance. The two ECG signals in the upper figure are close in the Wasserstein distance and similar to human perception while they have a large $L_1$ distance. In the bottom figure, the $L_{\infty}$ distance from the red curve to the blue and purple curve are both 2.88, but from human perception, the purple curve is much further away from the red curve than the blue curve. Therefore, the Wasserstein distance is more appropriate to measure the distance in this case.}
  \label{fig:intuition}
  \vspace{-0.4cm}
\end{figure}

However, recent studies have shown that adversarial examples can be generated to force a well-trained deep learning model to misclassify by applying small and unnoticeable perturbations to the input data \cite{szegedy2013intriguing}. The existence of such adversarial examples reveals the vulnerability of deep learning models, especially when applied to security-critical domains such as healthcare. 

The notion of indistinguishability of adversarial examples, in the context of computer vision, was originally taken to be $L_{\infty}$ bounded perturbations, which refers to noise with limited magnitude injected to each pixel \cite{goodfellow2014explaining}. Followed by that, the scale of adversarial perturbations is usually constrained by $L_p$ norm, %which is to search adversarial examples in, 
i.e., bounded by an $L_p$ norm ball centered at the normal example
\cite{kurakin2018adversarial, carlini2017towards}. As deep learning models are increasingly used for time series data, %and potential adversarial attacks are present in many applications where the use of time series data is crucial, 
several works  adapted  adversarial attacks from images to time series data \cite{TSC, AATS, MTSR} using $L_p$ norms. While adversarial examples under $L_p$ norm are intuitive for image classification tasks,
%as the Euclidean ball is a convenient source of adversarial perturbations on the image. 
$L_p$ norm is not an appropriate metric for the time series data.

In time series analysis, there are more effective metrics for measuring similarity between two temporal sequences, especially when the sequences vary in length and speed \cite{alt1995computing, berndt1994using}.  Wasserstein distance \cite{vallender1974calculation}, which is the numerical cost of an optimal transportation problem, allows us to analyze the distance between two time sequences. This distance can be intuitively understood for time sequence as the cost of moving around feature mass from one time step to another (transportation plan) in order to make two sequences the same. 
%%%%%%% remove %%%%%%
 \iffalse
The Wasserstein of two distribution $u, v \in \mathbb{R}^{n \times m}$ can be stated as:
\begin{equation*}
    d_W(u, v) = \underset{\Pi \in \mathbb{R}^{(n \cdot m) \times (n \cdot m)}}{min}< \Pi, \mathbf{C}>
\vspace{-0.3em}
\end{equation*}
where $\Pi$ refers to the transport plan and \mathbf{C} represents the cost of transporting a mass unit from each position.
\fi
 %%%%%%%% end remove %%%%%%
 
Note that Wasserstein distance and Euclidean distance are different measures and even opposite in some cases. Two time sequences can be close in the Wasserstein distance while far away from each other in Euclidean distance. As shown in Figure \ref{fig:intuition},  Wasserstein distance can better reflect human perception. The time series that appear more distinguished to our eyes are captured by larger Wasserstein distances. The example in the upper figure may not be a successful adversarial example under $L_1$ distance as it is distant in the $L_1$ space. However, it is almost imperceptible to human evaluation. In this case, Wasserstein distance is a better measurement of adversarial examples. 

In this work, we study the adversarial attack on time series in the Wasserstein space for the first time. % which better capture the distances. 
Our goal is to generate adversarial examples that have small Wasserstein perturbation so it is more indistinguishable and natural to human, e.g., physician who examines ECG data.
Projected gradient descent attack \cite{madry2017towards}  is a widely-used attack method that applies small steps of maximizing the loss objective iteratively and clipping the values of intermediate results after each step (projection to the $L_p$ norm ball) to ensure that they are in a constrained neighbourhood of the original inputs. Similarly, we propose a Wasserstein PGD method to search for adversarial examples in the Wasserstein space for univariant time series. Wasserstein distance cannot be calculated directly without solving an optimization subproblem and has no closed-form solution in most cases, which limits its applications. At present, there are only two cases that the Wasserstein distance can be directly calculated, one is the case of the dimension of inputs being 1, and the other is the inputs following Gaussian distribution. 
%In the paper\cite{wong2019wasserstein}, they applied Wasserstein PGD on the image classification task. As the projection cannot be directed calculated, they used Projected Sinkhorn Iterations to approximate the projection.
For the univariant time series, we can take advantage of its $1D$ characteristic and use the closed-form Wasserstein distance to apply the projection of intermediate results of each step onto the Wasserstein ball with gradient descent method.

\begin{figure}[htbp]
  \centering
  \includegraphics[width=0.5\textwidth]{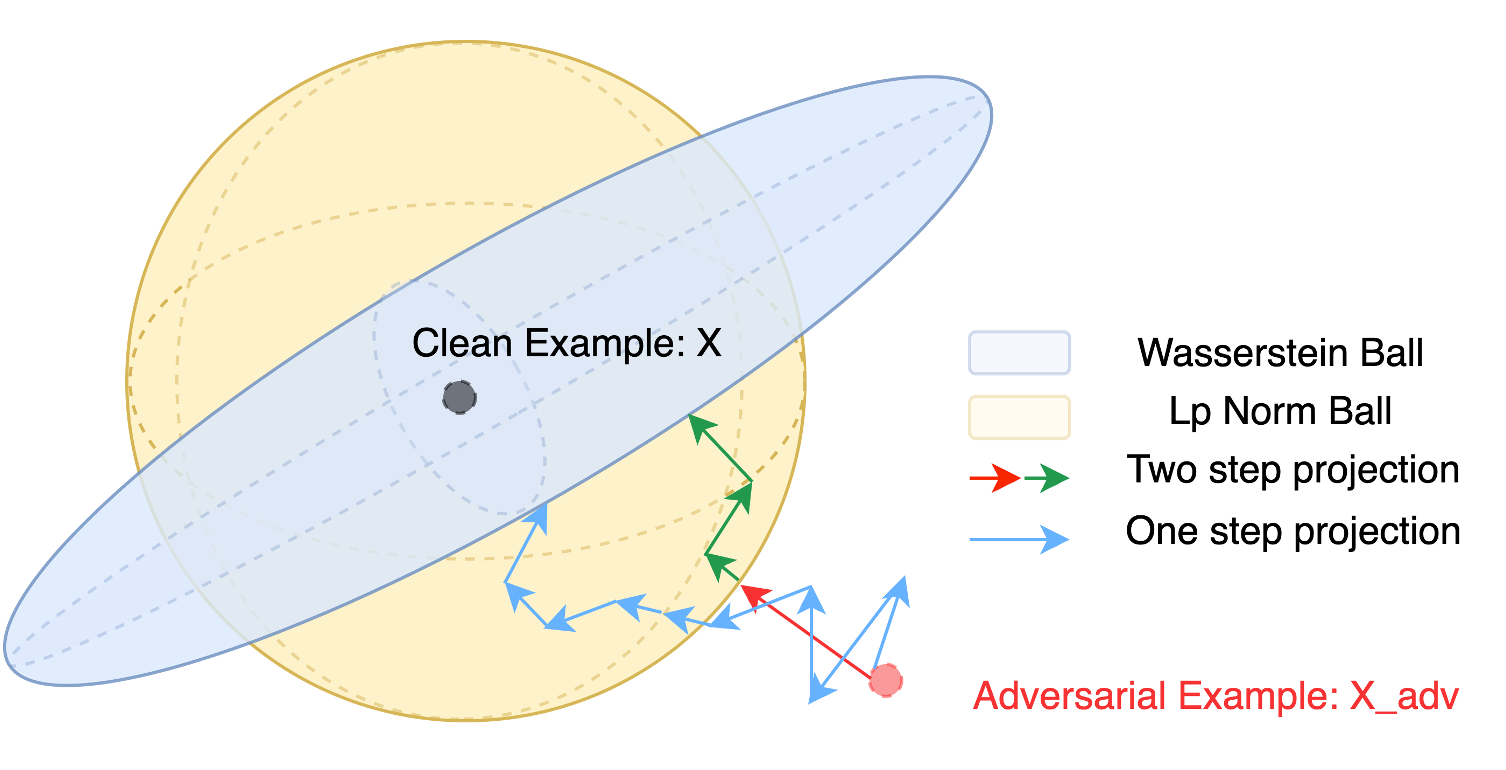}
  \caption{Illustration of the difference between direct projection and two-step projection}
  \label{fig:step}
%   \vspace{-0.4cm}
\end{figure}

%During experiments, we observed that 
The direct projection of intermediate results onto the Wasserstein ball using gradient descent could be time-consuming and converge to a sub-optimal solution. Thus, we further propose a two-step projection method that first projects the intermediate result to an $L_p$ norm ball (one step clipping) and then uses the projected example on the norm ball as the starting point for Wasserstein projection, so that the search in the Wasserstein space is guided and constrained. The intuition of the two-step projection is illustrated in Figure \ref{fig:step}, which will be explained in detail in the Method Section. 

The state-of-the-art defense mechanism to adversarial examples is the certified robustness approach which provides a theoretical guarantee that adversarial examples generated within certain distance bounds can be correctly classified. To better study the nature of Wasserstein adversarial examples, we also investigate how well the existing certified robustness approach work against our proposed Wasserstein attack. As Wasserstein adversarial examples are bounded by Wasserstein distance, the existing and well-known certified defense within Euclidean distance is not applicable \cite{cohen2019certified}. We adapt Wasserstein Smoothing \cite{levine2020wasserstein}, a certified robustness approach to Wasserstein adversarial examples for image data which transfers Wasserstein distance on the image into $L_1$ distance on the transport plan, to univariant time series adversarial examples. From the results, although the defense can achieve some accuracy gain, it still has limitations in many cases and leaves space for developing a stronger certified robustness method to Wasserstein adversarial examples on univariant time series data.

Our Contributions can be summarized as follows:

1) We study adversarial examples in the Wasserstein space for time series data for the first time which better capture the distance and are more natural and imperceptible.

2) We utilize the characteristics of univariant time series data and propose a projected gradient descent attack method 
which efficiently projects (bounds) adversarial examples in the Wasserstein ball.

3) We develop a two-step projection that first projects an adversarial example to an $L_p$ norm ball and then use the projected example as the starting point for Wasserstein projection to overcome the computing bottleneck of direct Wasserstein projection.

4) We empirically evaluate the proposed attack on several Electrocardiogram (ECG) datasets in the health care domain. Extensive results demonstrate that the Wasserstein PGD is powerful and can attack most of the target classifiers with a high attack success rate and yield more imperceptible and natural examples than attacks in the Euclidean space.

5) We evaluate Wasserstein smoothing designed for image data as a baseline certified robustness approach against Wasserstein attack which suggest that there is space for stronger defense mechanisms tailored to time series data.

%% file: related.tex
\vspace{-0.2cm}
\section{Related Work}
\vspace{-0.1cm}
In this section, we will first introduce  existing adversarial attacks in the image domain and time series domain. Followed by that we will discuss the state-of-the-art defense mechanism: certified robustness.
\vspace{-0.2cm}
\subsection{Adversarial Attack Methods}
\vspace{-0.1cm}
Adversarial examples were first introduced in the image domain \cite{szegedy2013intriguing}. Ian Goodfellow et al \cite{goodfellow2014explaining} proposed a fast Gradient Method to generate adversarial examples by performing a one-step gradient update along the direction of the sign of the gradient at each pixel. This method is simple and fast but cannot minimize perturbation. Other research attempted to iteratively apply FGSM in order to achieve a smaller perturbation such as I-FGSM and Projected Gradient Descent (PGD) \cite{madry2017towards, pWorld}. C\&W \cite{carlini2017towards} is a powerful attack that aims to both minimize the distance and maximizes classification loss  and hence achieves minimal perturbations for the same attack success rate. However, C\&W is relatively time-consuming. 

The above mentioned attack methods are all in the Euclidean space. \cite{wong2019wasserstein} opened up a new direction in adversarial examples in Wasserstein space for the image domain, by developing a procedure for projecting an example onto the Wasserstein ball. As the projection cannot be directed calculated, they used Projected Sinkhorn Iterations to approximate the projection. 

Several works have  adapted adversarial attacks from images to time series data. \cite{TSC} empirically studies the performance of I-FGSM in the time series classification tasks while \cite{MTSR} adopted I-FGSM into the time series regression tasks. \cite{AATS} utilized an adversarial transformation network (ATN) \cite{baluja2017adversarial} on a distilled model to attack various time series classification models. Although Wasserstein distance is a better distance measurement for time series data, no previous work has studied  adversarial examples in Wasserstein space for time series data. 

\vspace{-0.2cm}
\subsection{Certified Robustness}
\vspace{-0.1cm}
Certified robustness is the most powerful defense method to date as it is provable. The main idea is to transform the base classifier into a randomized classifier by applying random noise onto the inputs or classifier, such that the perturbed examples within a certain Euclidean ball are certified to have the same classification as the original example.  PixelDP\cite{PixelDP} utilized the definition of differential privacy (DP) \cite{mcsherry2007mechanism} to prove certified robustness using the Gaussian mechanism through noisy layers in the network. WordDP \cite{wang2021certified} extended the idea into the text domain using the Exponential mechanism. Randomized smoothing \cite{cohen2019certified} generates Gaussian noise on the input and provided a tighter certified bound using the Neyman Pearson lemma\cite{Neyman1992}.

Wasserstein smoothing \cite{levine2020wasserstein} was the first certified robustness method for adversarial images in Wasserstein space. The basic idea is to define a reduced transport plan and transfer Wasserstein distance on the image space to 
$L_1$ norm on the transport plan. Therefore, smoothing in the transport domain can be performed using  existing $L_1$ robustness certification provided by \cite{PixelDP} and transferred back to the Wasserstein space. In this paper, we will adopt Wasserstein smoothing
from the image domain to the univariant time series data and evaluate the performance of our proposed Wasserstein attack against this potential countermeasure.

%% file: method.tex
\vspace{-0.5em}
\section{Methods} 
In this section, we present our proposed Wasserstein adversarial example against time series models. We rely on the most common method of creating adversarial examples, the variation of projected gradient descent (PGD). The original PGD algorithm uses $L_\infty$ clipping to perform the projection while we will use Wasserstein projection in our method. First, we will explain how to perform the Wasserstein projection. Followed by that, we will explain the Wasserstein PGD algorithm and the two-step projection. 

%In the end, we briefly introduce the certified robustness mechanism we use for evaluating Wasserstein adversarial examples.

\vspace{-0.5em}
\subsection{Wasserstein Projection}
Let $(x, y)$ be a data point and its label, and $\mathcal{B}(x, \epsilon)$ be a ball around $x$ with radius $\epsilon$. The (general) projection of a point $w$ on to $\mathcal{B}(x, \epsilon)$ can be formulated as:
\begin{equation} 
    \underset{\mathcal{B}(x, \epsilon)}{proj(w)} = \underset{z \in \mathcal{B}(x, \epsilon)}{\argmin} ||w-z||_2^2.
\end{equation}
A Wasserstein ball around sample $x$ with radius $\epsilon$ can be defined as:
\begin{equation} 
    \mathcal{B}_w(x, \epsilon)= \{x+\delta: d_\mathcal{W}(x, x+\delta) \leq \epsilon\},
\end{equation}
where $d_\mathcal{W}(u, v)$ refers to the Wasserstein distance between two sample distributions  in the space $\mathcal{X} = \mathbb{R}^2$, which can be calculated as:
\begin{equation} 
    d_\mathcal{W} (u, v) = [\underset{\gamma \in \Pi(u, v)}{inf} \int_{\mathcal{X}^2} ||x-y||^p d\gamma(x, y)]^{1/p}.
\end{equation}
Here $\gamma$ defines the joint probability distribution, called coupling, which has marginal distribution exactly as $u$ and $v$.
\vspace{-0.1cm}
\begin{algorithm}
\caption{1D\_Wasserstein\_Projection}\label{alg:projection}
\KwIn{
     Original input: $x$; Initial adversarial input $x_{adv}^0$; Wasserstein projection bound: $\mathcal{C}$;
     Step size: $\alpha$}
\KwOut {
    Projected Adversarial example: $x_{adv}$}
 $i \gets 0$\;
 $L_{w}^0 =d_\mathcal{W}(x, x_{adv}^0)$ (Formula \ref{equ:dw})\;
\While{$L_{w} \geq \mathcal{C}$}
{
         $x_{adv}^{i+1} = x_{adv}^i - \alpha \cdot \frac{\partial L_{w}(x, x_{adv}^i) }{\partial x_{adv}^i}$\;
         $L_{w}^i = d_\mathcal{W}(x, x_{adv}^i)$\;
         i++\;
        \If {$i \geq \mathbb{I}$}
        {
             Break\;
             \textbf{Return} False\;
        }
}
\end{algorithm}
\vspace{-0.1cm}

When the input distribution satisfies the dimension being one, there is a closed-form solution for the above $d_\mathcal{W}$:
\begin{equation} \label{duv}
\begin{split}
d_\mathcal{W} (u, v) & = ||F_u^{-1} - F_v^{-1}||^p \\
                     & = (\int_{0}^{1} ||F_u(\alpha)^{-1}- F_v(\alpha) ^{-1}||^p d \alpha)^{1/p}.
\end{split}
\end{equation}
When $p$ equals 1 and the inputs are in the discrete case, formula \ref{duv} can be further simplified as:
\begin{equation}
\begin{split}
     d_\mathcal{W} (u, v) & = \int_{\mathbb{R}}| F_u(\alpha)- F_v(\alpha) | d \alpha \\
                            & = \sum_{i = 1}^n | \sum_{j = 1}^i u_i - \sum_{j = 1}^i v_i|.
\label{equ:dw}
\end{split}
\end{equation}
In this case, the transport plan is $t = F_v^{-1} \odot F_u$.

Specifically, projecting $w$ onto the Wasserstein ball around $x$ with radius $\epsilon$ is defined as:
\begin{equation} 
    \underset{\mathcal{B_\mathcal{W}}(x, \epsilon)}{proj(w)} = \underset{z \in \mathcal{B}_\mathcal{W}(x, \epsilon)}{\argmin} d_\mathcal{W} (x, z).
\end{equation}
As $1D$ Wasserstein distance has a closed-form solution and the above formula is also differentiable, we can apply the gradient descend method to the projection, the algorithm is explained in Algorithm \ref{alg:projection}. Note that this method may not find the exact projection onto the Wasserstein ball, but it can converge to an example within the Wasserstein ball. %In the original PGD algorithm, the author uses $L_\infty$ clipping to perform the projection.

\subsection{Wasserstein PGD Attack}
\vspace{-0.1cm}
\begin{algorithm}

\caption{ Wasserstein PGD Attack on Univarient Time series}\label{alg:cap}
\KwIn{
     Original input: $\mathcal{X} = \{x, y\}$; Target model: $\mathcal{F_{\theta}}$; Attack Iteration: $\mathbb{T}$; Step size: $\epsilon$; Wasserstein projection bound: $\zeta$; $L_{\infty}$ Norm clipping value: $\delta$ \\}
\KwOut {
    Adversarial example: $x_{adv}$}
    $S_t \gets 0$\;
 Init $x_{adv}^0 = x + \mathcal{N}(0,1)$\;
\While{$t \leq \mathbb{T}$}
{       \tcc{gradient descend step}
        $\eta = \epsilon \cdot sign(\nabla_x L_{cross\_entropy}(x, y, \mathcal{F}_{\theta}))$\;

         $x_{adv}^{t+1} = x_{adv}^t +\eta$\;
          \tcc{norm ball clipping with center $x$ and radius $\zeta$}
          $x_{adv}^{t+1} = \min(\max(x_{adv}^{t+1}, x_{adv}^{t+1}-\zeta), x_{adv}^{t+1}+\zeta)$\; 
          
          \tcc{Wasserstein ball projection with center $x$ and radius $\delta$}
         $x_{adv}^{t+1}$ = 1D\_Wasserstein\_Projection($x, x_{adv}^{t+1}, \delta)$ (Algorithm \ref{alg:projection})\;

         t++\;
}
\end{algorithm}
\vspace{-0.1cm}

PGD attack utilizes the concept of back-propagation. It takes the gradient of the loss function over inputs, to generate small perturbations at each time step and iteratively adds the perturbation to the clean input to generate adversarial examples followed by projection (clipping into the $L_\infty$ ball), such that the new adversarial example has a greater tendency towards being misclassified. For Wasserstein adversarial attack, each iteration of the algorithm includes a gradient descent step to update the perturbed example followed by Wassersten projection, which is formulated as:

\begin{equation}
    x^{t+1}_{adv} = \underset{\mathcal{B}_\mathcal{W}(x, \epsilon)}{proj}  (x^t_{adv} +  \alpha ^T \nabla \mathcal{L}(x^t_{adv}, y)).
\end{equation}

Theoretically, Wasserstein projection can be performed from any starting point. However, %during experiments, we noticed that 
direct projection onto the Wasserstein ball using gradient descent can be time consuming and converge to a sub-optimal solution. We propose a two-step projection method (shown in Algorithm 2) that first projects the adversarial example to a norm-ball and then uses the projected example as the starting point for Wasserstein projection. As shown in Figure \ref{fig:step}, the blue curve refers to the direct Wasserstein projection using gradient descent (this is the gradient descent used to perform projection shown in line 3-5 in Algorithm 1). This is performed after each  step of gradient descent that updates the adversarial example (this is the gradient descent used to generate the perturbation shown in line 4-5 in Algorithm 2). The red curve and green curve refer to the two-step projection that first projects (clips) the example to the norm ball (line 6) and then projects it to the Wasserstein ball (line 7). In this way, the search in the Wasserstein space is guided and constrained, which is more effective and efficient.

\iffalse
$x_{adv}: v$\\
$x_{adv}$ after projection: $v'$\\
$x_{clean}: u$\\

\begin{equation}
   d_\mathcal{W} (u, v) & = \int_{\mathbb{R}}| F_u(\alpha)- F_v(\alpha) | d \alpha  
\end{equation}

we want to project v into a Wasserstein ball with center $u$ and radius $t$:
\begin{equation*}
\begin{split}
    
   &min \int_{\mathbb{R}}| F_{v'}(\alpha)- F_v(\alpha) | d \alpha \\
   
   &st. \int_{\mathbb{R}}| F_u(\alpha)- F_v(\alpha) | d \alpha  \leq t
  \end{split}
\end{equation*}

\fi

%% file: results.tex
\vspace{-0.2cm}
\section{Experiments}
In this section, we will first describe our experimental settings and then evaluate our proposed Wasserstein PGD in terms of the Attack Success Rate (ASR, the fraction of examples that label has been fliped), and comparison with the PGD attack in the Euclidean space. We also demonstrate the effectiveness of the 2-step projection. Finally, we will demonstrate the results of certified robustness to the proposed Wasserstein PGD attack.
\begin{table}[h]
\centering
\begin{tabular}{|l|l|l|l|l|} 
\toprule
\hline
~ &\multicolumn{4}{l|}{\textbf{Data Description}} \\
\hline
\textbf{Dataset} & \textbf{TrainSize}          & \textbf{TestSize}    & \textbf{Classes} & \textbf{SeqLen}   \\ 
\hline
\textbf{ECG200}  & 100        & 100     &2 &96\\
\hline
\textbf{ECG5000}  & 500        & 4500     &5 &160  \\
\hline
\textbf{ECGFiveDays}  & 23        & 861     &2 &136 \\
\hline
~ &\multicolumn{4}{l|}{\textbf{Model Performance}} \\
\hline
 \textbf{Dataset} & \textbf{MLP}   & \textbf{FCN}  &\textbf{CNN}   &\textbf{ResNet} \\
\hline
\textbf{ECG200} &0.916 &0.9 &0.83 &0.89 \\
\hline
\textbf{ECG5000} &0.931 &0.939 &0.928 &0.934 \\
\hline
\textbf{ECGFiveDays} &0.979 &0.987 &0.885 &0.993 \\
\hline
\end{tabular}
\caption{Summary of datasets}
\label{tab: summary}

\end{table}
\vspace{-0.1cm}

\begin{figure*}[ht]
 \centering
 %%%%%%%% remove %%%%%%
     \begin{subfigure}[b]{0.33\textwidth}
         \centering
         \includegraphics[width=\textwidth]{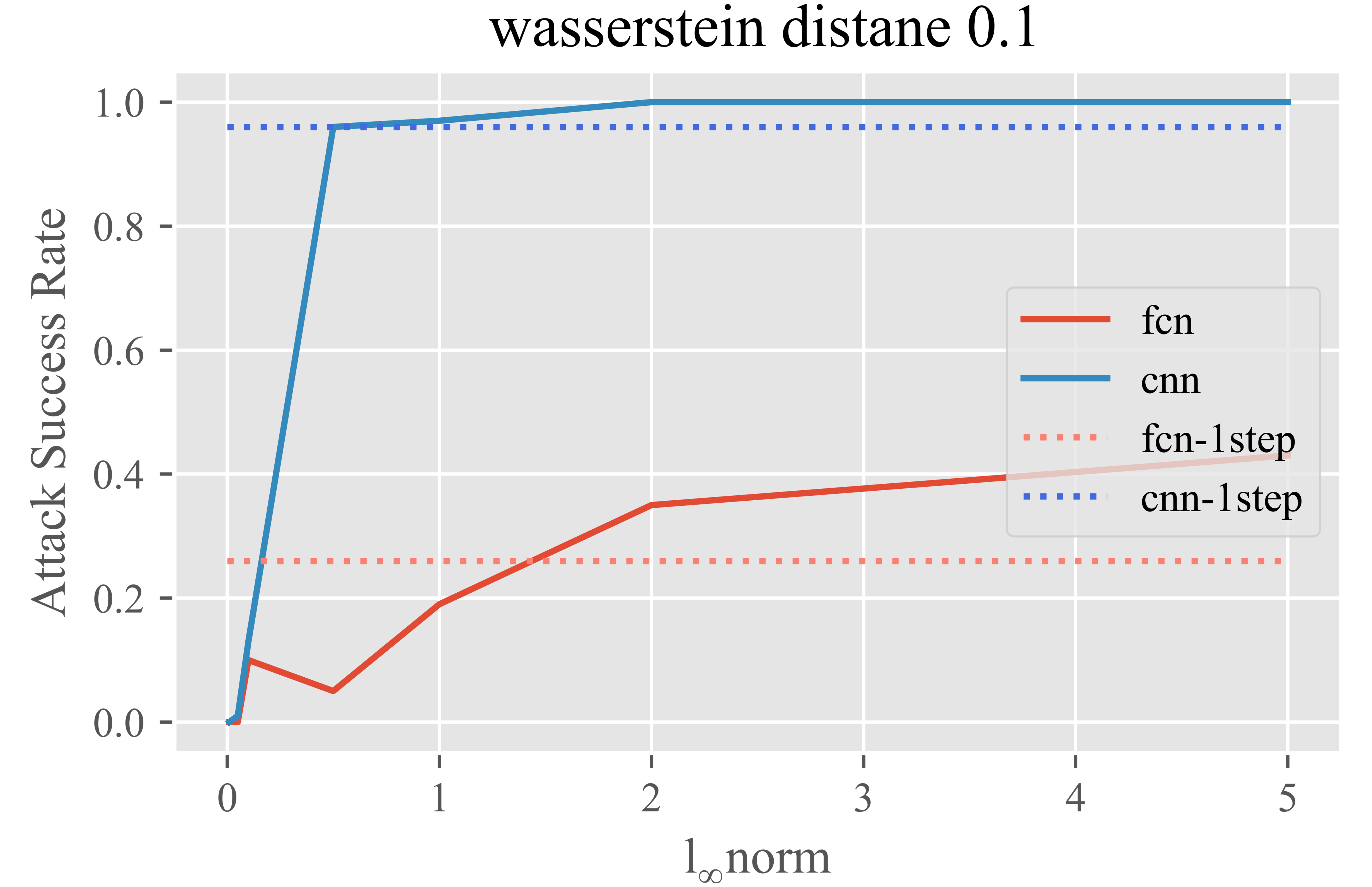}
         \label{fig:y equals x}
     \end{subfigure}
     \begin{subfigure}[b]{0.33\textwidth}
         \centering
         \includegraphics[width=\textwidth]{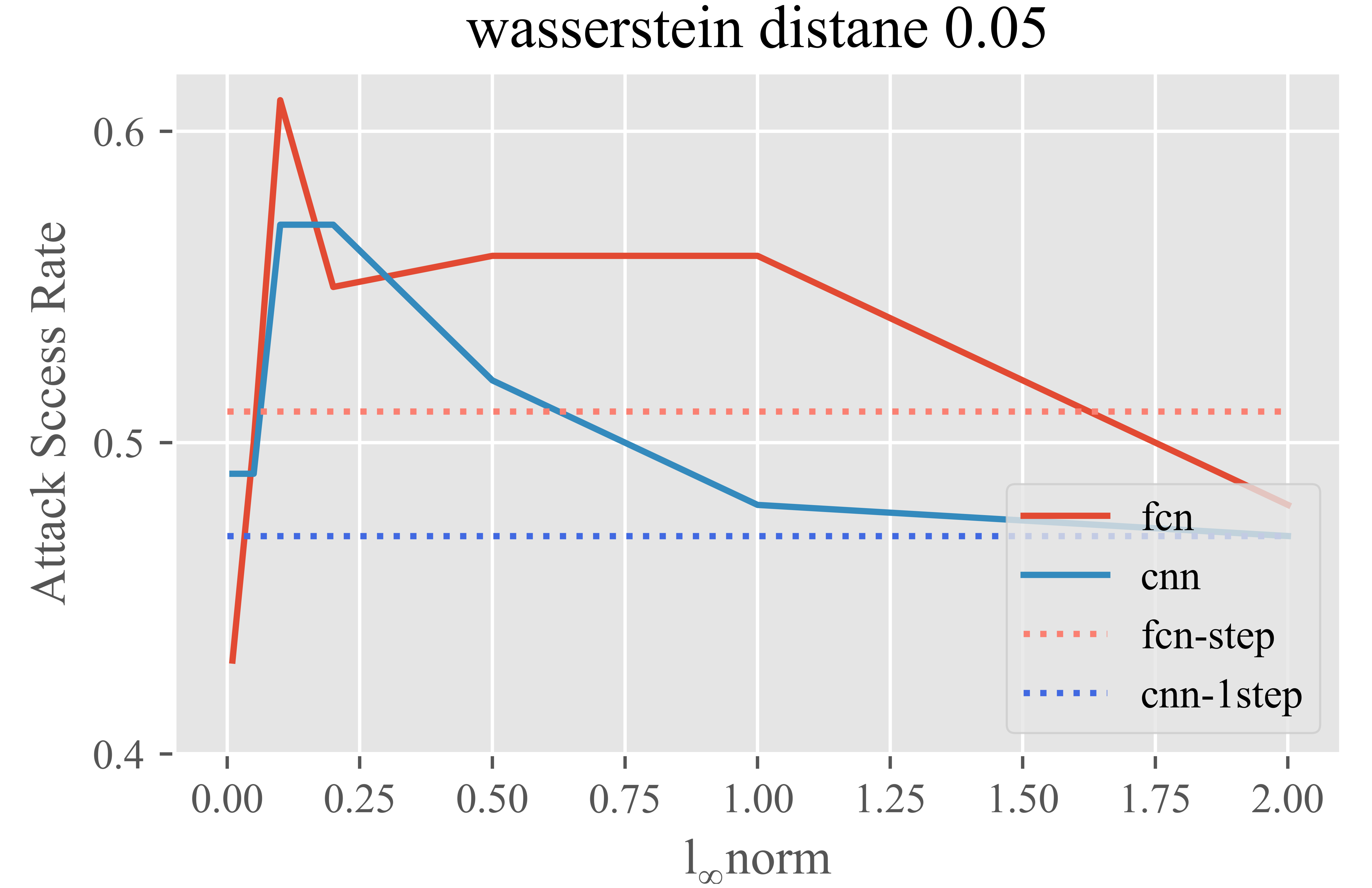}
         \label{fig:three sin x}
     \end{subfigure}
     \begin{subfigure}[b]{0.33\textwidth}
         \centering
         \includegraphics[width=\textwidth]{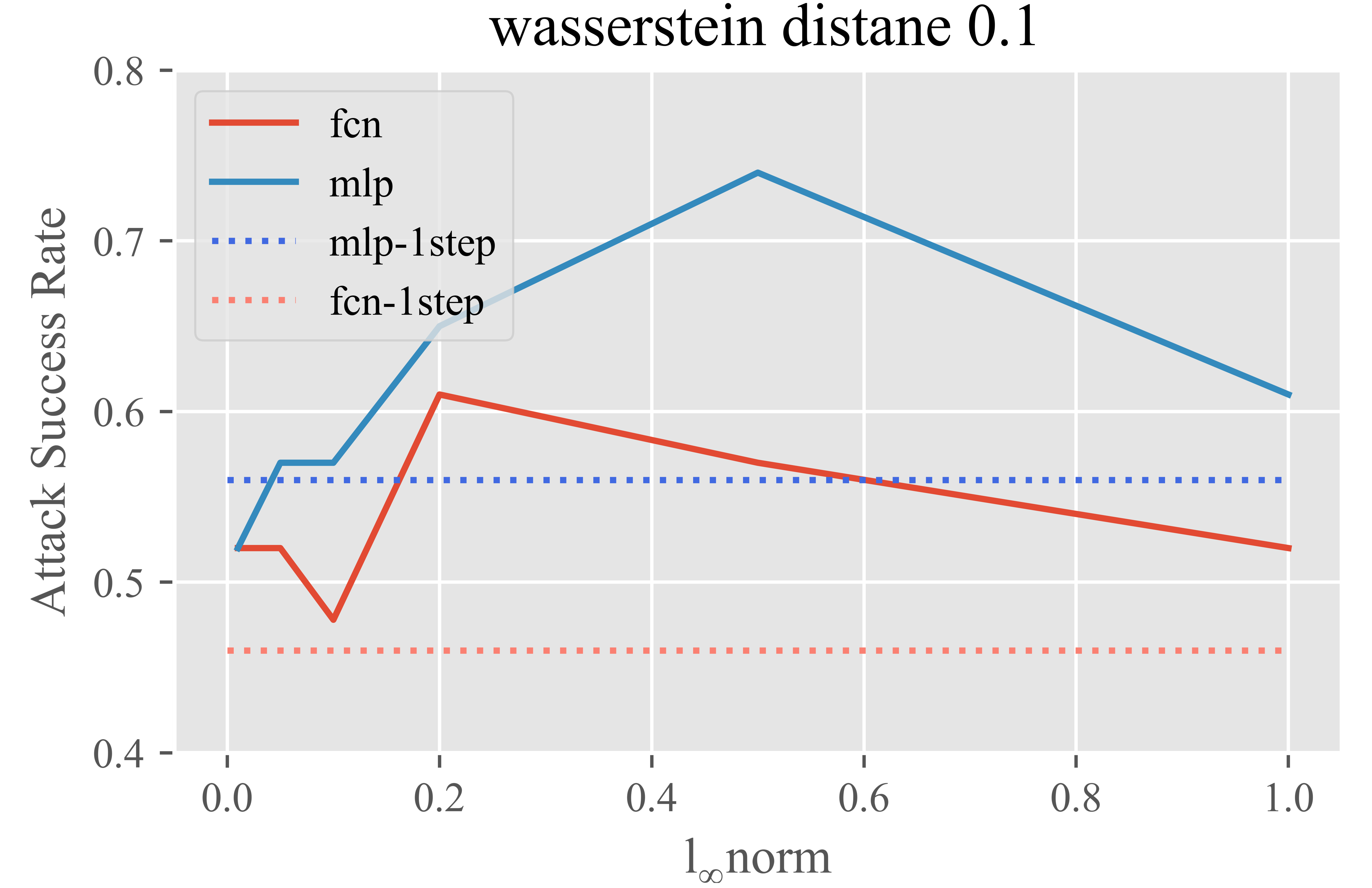}
         \label{fig:five over x}
     \end{subfigure}
     \begin{subfigure}[b]{0.33\textwidth}
         \centering
         \includegraphics[width=\textwidth]{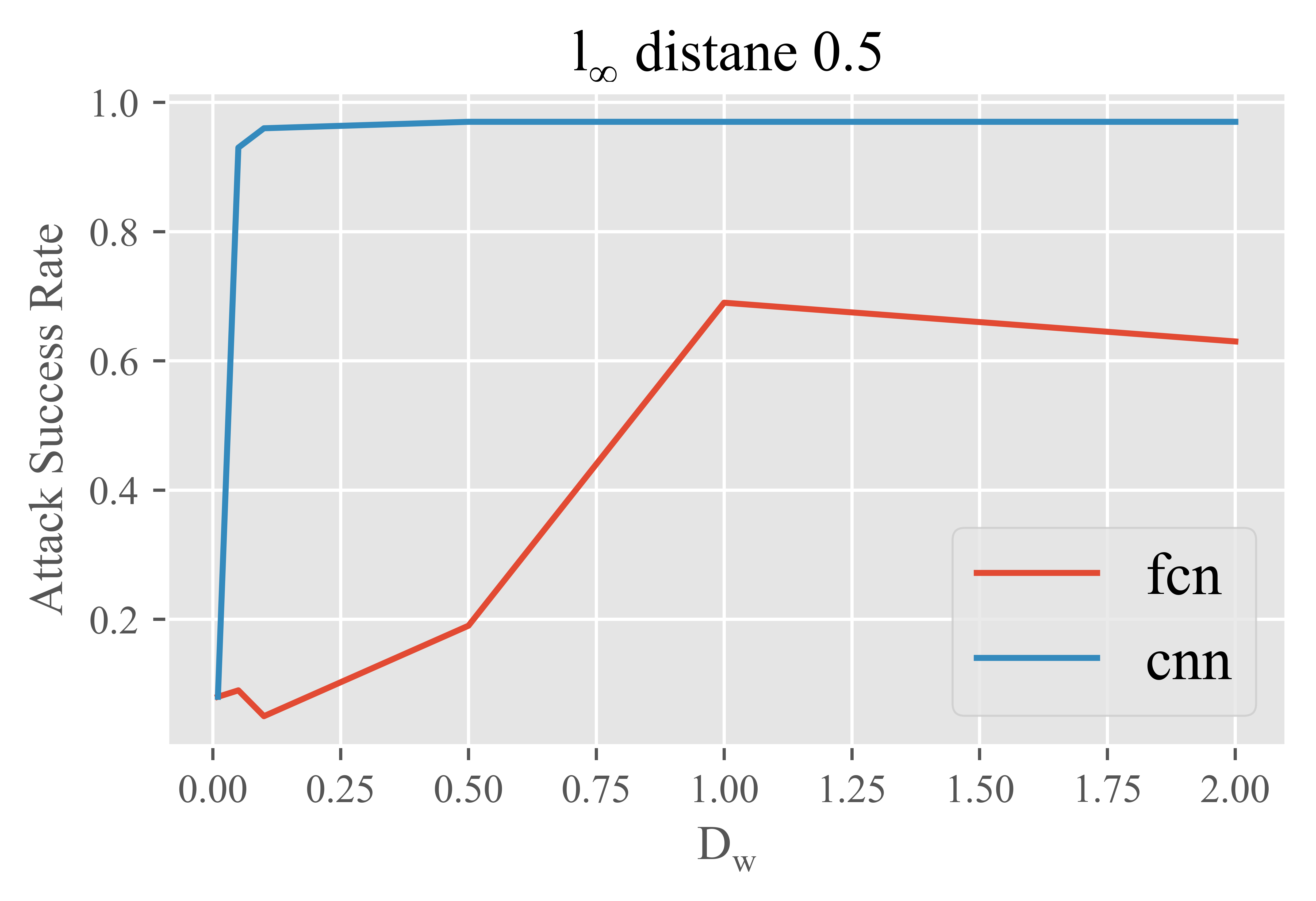}
         \caption{ECG5000}
         \label{fig:y equals x}
     \end{subfigure}
     \begin{subfigure}[b]{0.33\textwidth}
         \centering
         \includegraphics[width=\textwidth]{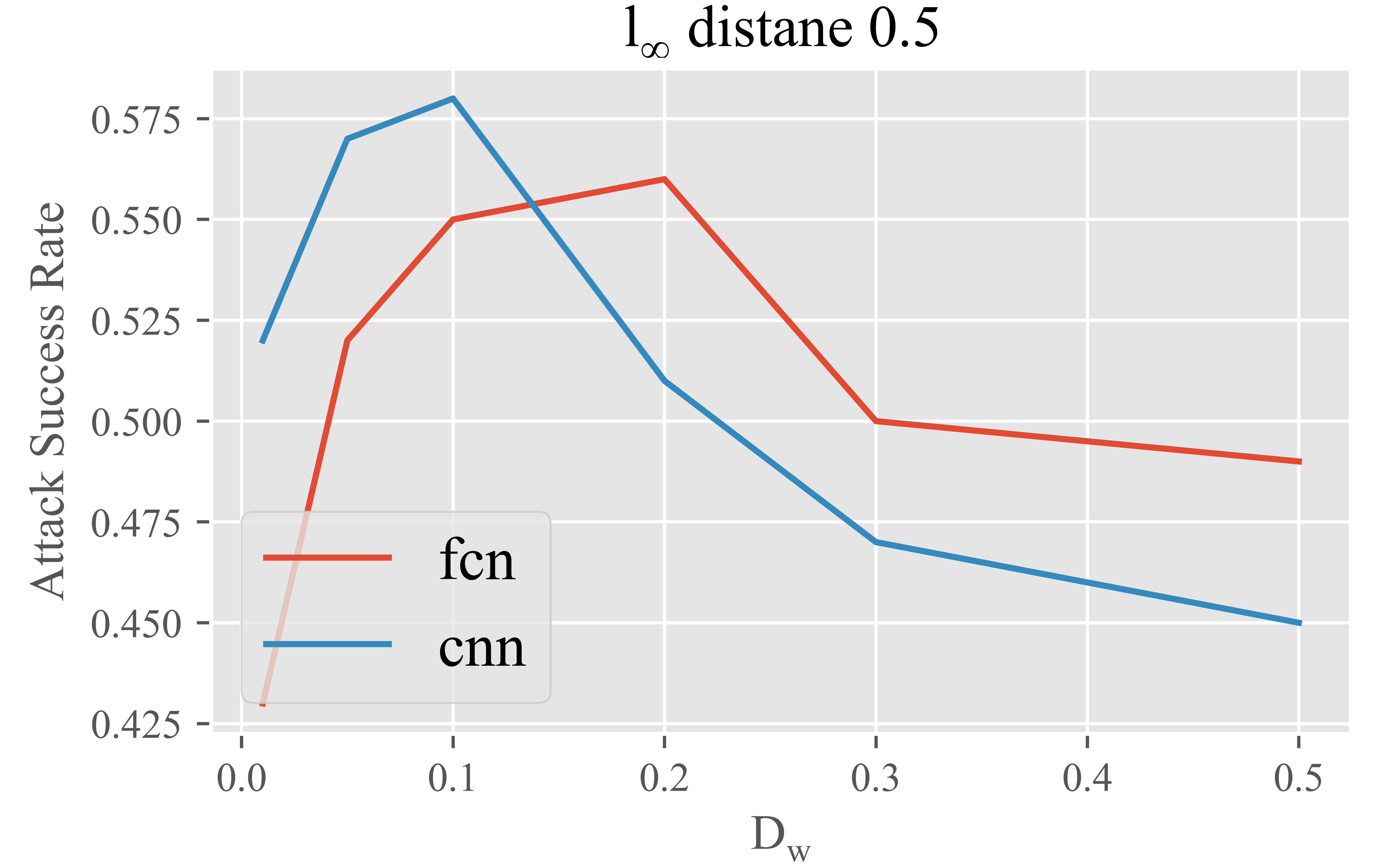}
         \caption{ECG200}
         \label{fig:three sin x}
     \end{subfigure}
     \begin{subfigure}[b]{0.33\textwidth}
         \centering
         \includegraphics[width=\textwidth]{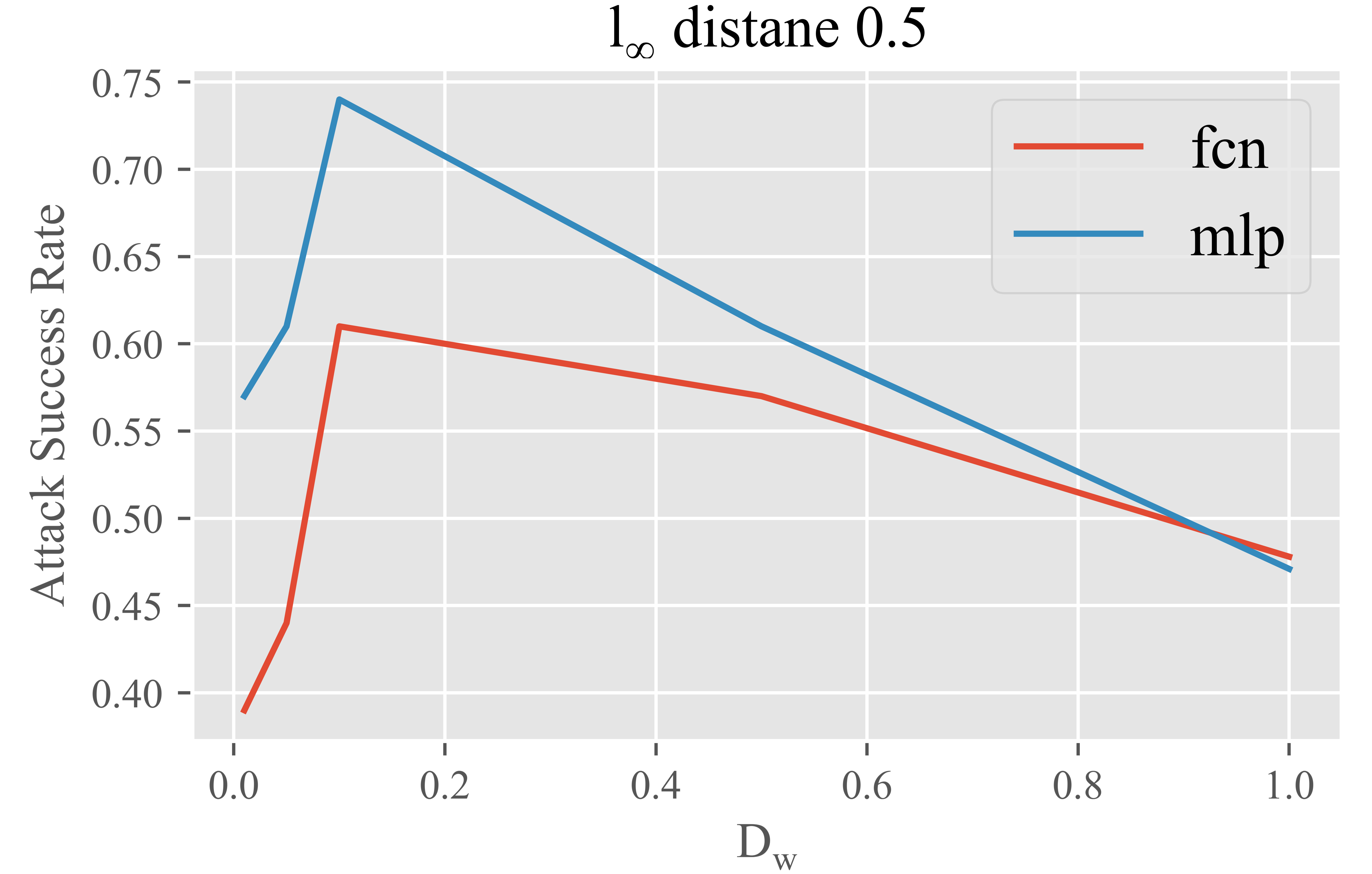}
         \caption{ECGFiveDays}
         \label{fig:five over x}
     \end{subfigure}
 %%%%%%%% end remove %%%%%%
  \iffalse
    \begin{subfigure}[b]{0.33\textwidth}
         \centering
         \includegraphics[width=\textwidth]{./fig/ECG5000.png}
         \caption{ECG5000}
         \label{fig:y equals x}
     \end{subfigure}
     \begin{subfigure}[b]{0.33\textwidth}
         \centering
         \includegraphics[width=\textwidth]{./fig/ECG200.png}
         \caption{ECG200}
         \label{fig:three sin x}
     \end{subfigure}
     \begin{subfigure}[b]{0.33\textwidth}
         \centering
         \includegraphics[width=\textwidth]{./fig/ECGFD.png}
         \caption{ECGFiveDays}
         \label{fig:five over x}
     \end{subfigure}
\fi
\caption{Attack Success Rate under different $\l_{\infty}$ and Wasserstein Bound: The columns represent the three dataset respectively.The first row illustrate under the same Wasserstein distance bound, how the attack success rate change with the increase of the $\l_{\infty}$ bound; The Second row illustrate under the same $\l_{\infty}$ bound, how the attack success rate change with the increase of the Wasserstein distance bound.}
\label{fig:three graphs}
\end{figure*}

%%%%%%%%%%%%%%%%%%%%%%%%%%%%%%%%%%%%%%%%%%%%%%%%
%%%%%%%%%%%%%%%%%%%%%%%%%%%%%%%%%%%%%%%%%%%%%%%%
%%%%%%%%%%%%%%%%%%%%%%%%%%%%%%%%%%%%%%%%%%%%%%%%
\vspace{-0.5em}
\subsection{Experimental Setup}
\vspace{-0.3em}
Our experiments are evaluated on Five benchmark time series classification datasets from the publicly available UCR archive \cite{UCRArchive}. The datasets are selected under the ``ECG" category for ECG based diagnosis tasks, where an adversarial attack is a potential security concern. In this section, we only show the results on three of the five ECG dataset, which are: 1) ECG200 which includes two classes (normal heartbeat and  Myocardial Infarction), and contains 35 half-hour records sampled with the rate of 125 Hz. 2) ECG5000 which is the Beth Israel Deaconess Medical Center (BIDMC) congestive heart failure database, consisting of records of 15 subjects, with severe congestive heart failure. Five labels refes to different levels of heart failure. Records of each individual were recorded in 20 hours, containing two ECG signals, sampled with the rate of 250 Hz. 3) ECGFiveDays which is from a 67-year-old male, including two classes which are two ECG dates. We relegate the full version of the experiments of the other two datasets to the Appendix.

We adopt and evaluate the target deep learning models from \cite{ismail2019deep} including: Multi-Layer Perceptron (MLP) \cite{gardner1998artificial}, Fully Connected Networks (FCN)\cite{long2015fully}, Convolutional Neural Networks (CNN) \cite{krizhevsky2012imagenet} and Residual Networks (ResNet) \cite{he2016deep}. The detailed information of the ECG datasets and the performance of the target models on each dataset is listed in Table \ref{tab: summary}.

%%%%%%%%%%%%%%%%%%%%%%%%%%%%%%%%%%%%%%%%%%%%%%%%
%%%%%%%%%%%%%%%%%%%%%%%%%%%%%%%%%%%%%%%%%%%%%%%%
%%%%%%%%%%%%%%%%%%%%%%%%%%%%%%%%%%%%%%%%%%%%%%%%

\vspace{-0.2cm}
\subsection{Attack Success Rate}
\vspace{-0.4em}
Our proposed 2-step projection Wasserstein PGD involves a first projection to a $L_{\infty}$ norm-ball and a second projection to a Wasserstein ball. Figure \ref{fig:three graphs} illustrates the impact of these two projections by comparing the ASR under different $L_{\infty}$ and Wasserstein bounds. The first row shows under the same Wasserstein distance bound, how the ASR changes with the increase of the $L_{\infty}$ bound, while the Second row shows under the same $\L_{\infty}$ bound, how the ASR changes with the increase of the Wasserstein distance bound. Each column represents the results of each dataset and each line in each figure corresponds to a target model. We show two models for each dataset and relegate the full version of the experiments to the Appendix. 

%For ECG5000 dataset, 
Under the same radius of Wasserstein ball, the general trend is that ASR first increases with the increase of $L_{\infty}$ bound. This is intuitive as the search space for optimal adversarial examples that satisfy the Wasserstein distance constraint is increasing. It is easier to find an adversarial example that successfully attack the target model and meanwhile satisfy the Wasserstein constraint. However, as the $L_{\infty}$ bound keeps increasing, it does not help anymore and even hurts the performance because the search in the Wasserstein Space is not guided and constrained any more and can be too large for the projection to find a good solution.   Therefore, the ASR stops increasing or starts to decrease.  This decreasing trend is more noticeable in ECG200 and ECGFiveDays datasets, while for ECG5000, the ASR increases to and stays at 1 (for the CNN model). Note that our attack is untargeted attack that aims to flip the label to any class other than the original label rather than the targeted label. Therefore, multi-class classification task (ECG5000) is easier to attack than binary classification tasks (ECG200 and ECGFiveDays). 

\begin{figure}[ht]
\centering
  \includegraphics[width=0.5\textwidth]{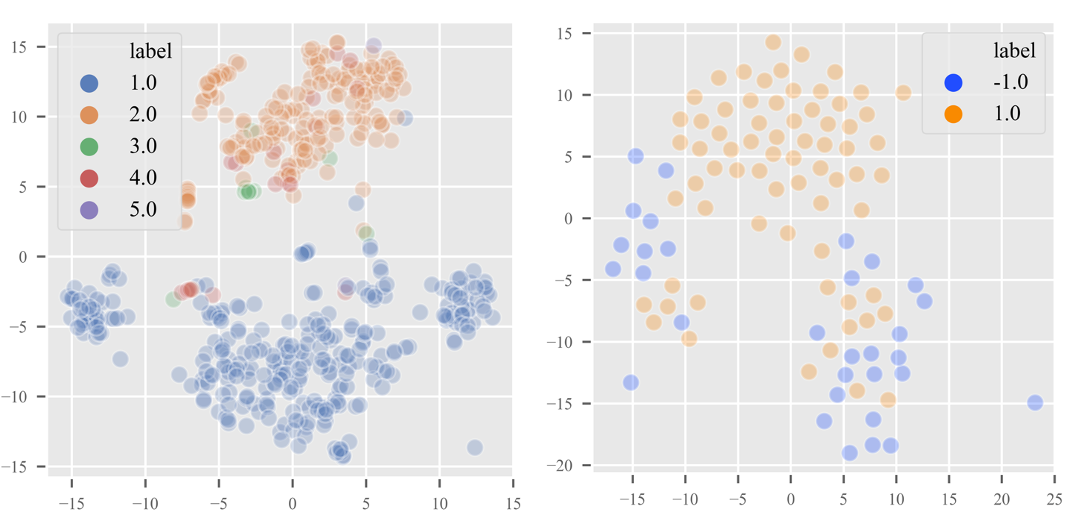}
  \caption{t-sne for ECG5000 (left) and ECG200 (right)}
  \label{fig:tsne}
 \vspace{-0.4cm}
\end{figure}

To further explain the difference between the datasets, we use t-distributed stochastic neighbor embedding (t-SNE), a nonlinear dimensionality reduction technique for high-dimensional visualization in a low-dimensional space \cite{JMLR:v9:vandermaaten08a}, to visualize the datasets. Figure \ref{fig:tsne} shows the 2D t-sne for ECG200 and ECG5000 respectively. Each color represents a class label. We can note that data points of ECG200 is more separable and the class boundary is more clear, while the classes are overlapping for ECG5000 and the boundary is less clear, which makes it easier to attack. This explains why for ECG5000, the ASR increases to 1 and does not decrease.

On the other hand, under the same $L_{\infty}$ bound shown in the bottom row of Figure \ref{fig:three graphs}, larger Wasserstein bound  also renders higher ASR at first due to larger search space. As it keeps increasing, the ASR decreases due to the search space being too large and the ineffectiveness of the search, especially when the radius of Wasserstein ball is greater than the Euclidean ball. Overall, the ASR of ECG5000, ECG200 and ECGFivedays can reach 100\%, 62\% and 74\% respectively.

%%%%%%%%%%%%%%%%%%%%%%%%%%%%%%%%%%%%%%%%%%%%%%%%
%%%%%%%%%%%%%%%%%%%%%%%%%%%%%%%%%%%%%%%%%%%%%%%%
%%%%%%%%%%%%%%%%%%%%%%%%%%%%%%%%%%%%%%%%%%%%%%%%
\vspace{-0.4em}
\subsection{Effectiveness of 2-step Projection}
\vspace{-0.4em}

From the perspective of ASR, we compare the 2-step projection with the direct 1-step Wasserstein projection shown as the dotted lines in the top row of Figure \ref{fig:three graphs} under the same attack settings. We observe that 2-step projection can achieve a higher ASR in general and optimal attack success rate when choosing the proper $L_{\infty}$ bound for the first projection.

\begin{figure}[ht]
\vspace{-0.2cm}
  \setlength{\abovecaptionskip}{0.cm}
  \centering
  \includegraphics[width=0.42\textwidth]{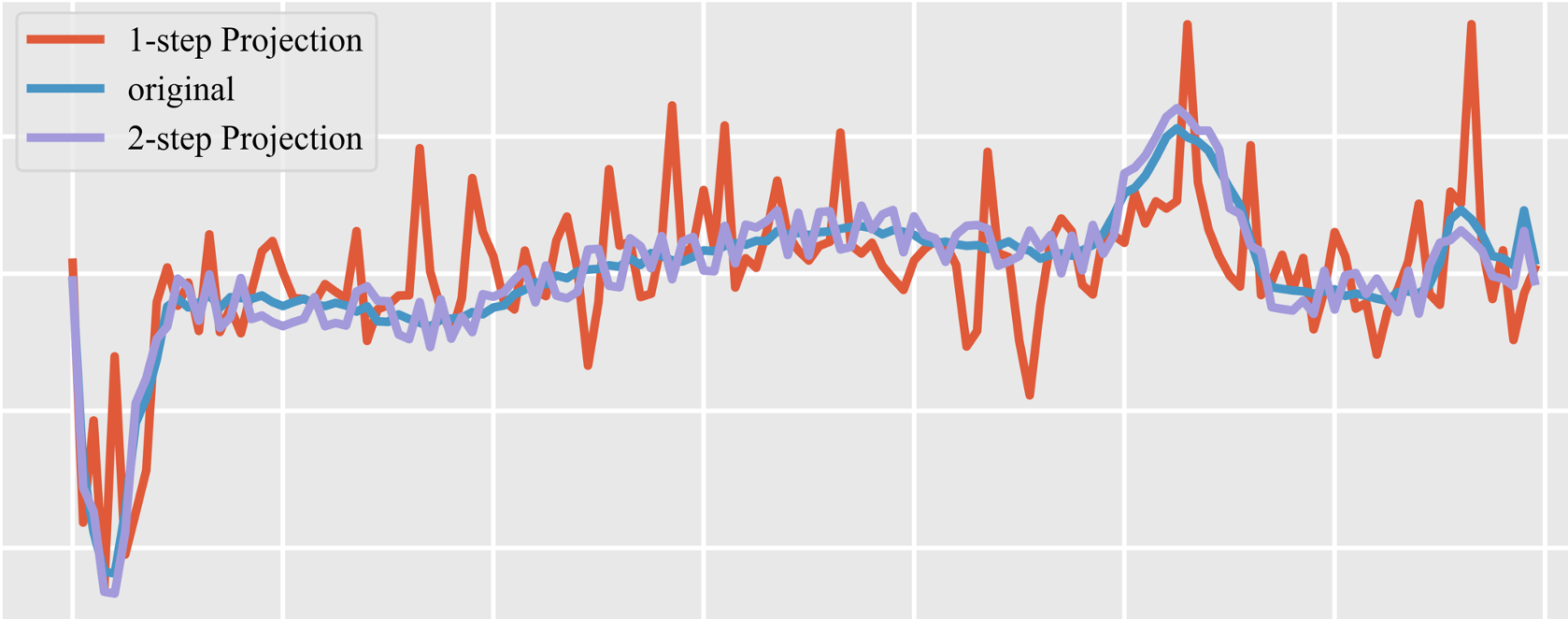}
  \caption{Comparison between direct Wasserstein projection (1-step projection) and 2-step projection.}
  \label{fig:1step-2step}
  \vspace{-0.1cm}
\end{figure}

From the perspective of human inspection,  Figure \ref{fig:1step-2step} shows two successful adversarial examples generated by the 2-step projection (purple) and direct  projection (red) respectively in comparison with the original example (blue). Although the Wasserstein perturbation distances of the two adversarial examples are both 0.99, the one that is first norm clipped is more imperceptible to human eyes, which has not only small Wasserstein distance but also bounded by $L_{\infty}$ distance. 
\vspace{-0.2cm}

\subsection{Comparison with $L_{\infty}$ PGD}
\setlength{\abovecaptionskip}{0.cm}
%%%%%%%%%%%%%%%%%%%%%%%%%%%%%%%%%%%%%%%%%%%%%%%%
%%%%%%%%%%%%%%%%%%%%%%%%%%%%%%%%%%%%%%%%%%%%%%%%
%%%%%%%%%%%%%%%%%%%%%%%%%%%%%%%%%%%%%%%%%%%%%%%%
%\vspace{-0.2cm}

The intuition of developing the Wasserstein PGD attack is to search for more indistinguishable and natural adversarial examples in the Wasserstein space. Therefore, we compare the adversarial examples generated by Wasserstein PGD with those generated by original PGD in the Euclidean space ($L_{\infty}$ PGD). On one hand, We draw the utility comparison from two aspects: 1) Under the same attack success rate, Wasserstein PGD is more natural; and 2) Under the same perturbation scale, Wasserstein PGD has a higher attack success rate. On the other hand, as Wasserstein projection involves gradient descent which will add more time cost in generating adversarial examples, we also compare the average time cost of two attack methods.

\begin{figure}[ht]
\vspace{-0.2cm}
  \centering
  \includegraphics[width=0.45\textwidth]{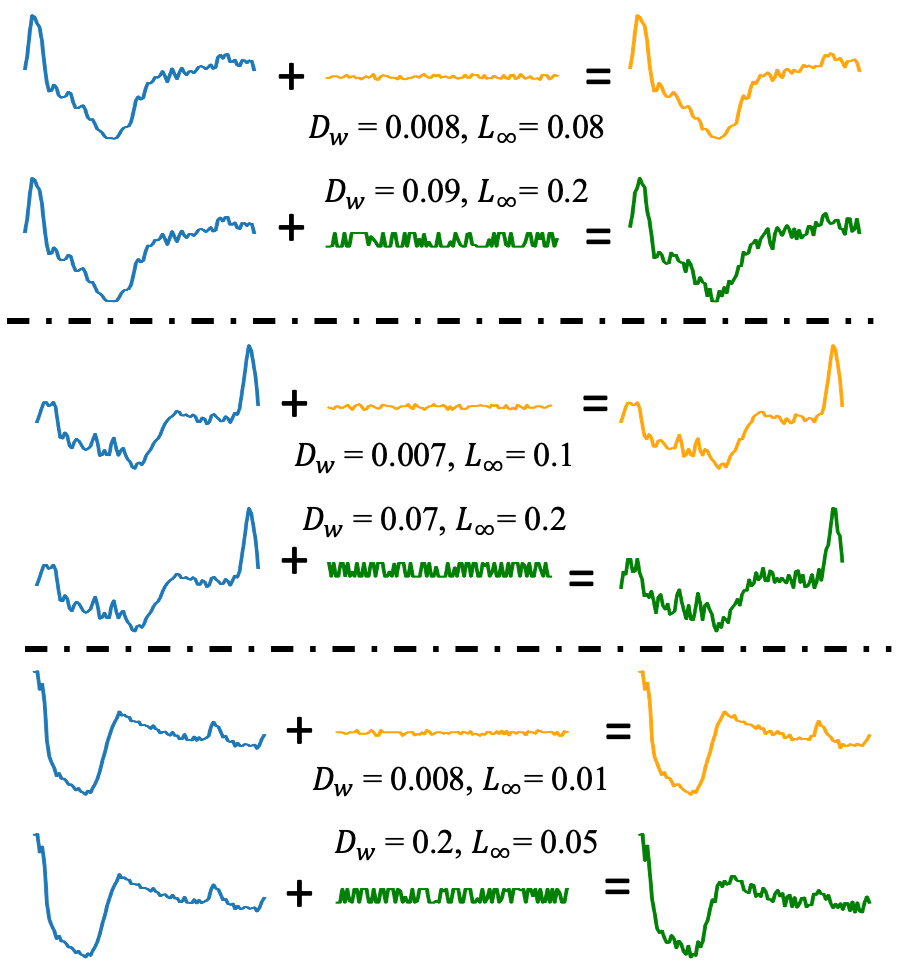}
  \caption{Comparison between Wasserstein PGD (yellow) and $L_{\infty}$ PGD (green) under the same attack success rate.}
  \label{fig:compare}
 \vspace{-0.1cm}

\end{figure}
\vspace{-0.1cm}
\subsubsection{Under the same attack success rate, Wasserstein PGD is more natural}
Figure \ref{fig:compare} illustrates several comparisons between Wasserstein PGD and $L_{\infty}$ PGD under the same ASR. We selected three examples randomly. For each figure, the blue curve is the original input. The yellow curves represent the perturbation and adversarial example generated by Wasserstein PGD, while the green curves represent the $L_{\infty}$ PGD. Clearly, the perturbation generated from Wasserstein PGD is smaller and more indistinguishable than $L_{\infty}$ PGD. 

\subsubsection{Under the same perturbation scale, Wasserstein PGD has a higher attack success rate.}
Another aspect to show the effectiveness of Wasserstein PGD is to compare the ASR with the original PGD under the same perturbation scale. However, the two attacks are conducted in different spaces. It is unfair to compare the ASR of adversarial examples generated from the Wasserstein ball and the $L_{\infty}$ ball with the same radius, as they represent completely different spaces.
To overcome this challenge, we use the greatest $L_{\infty}$ norm of the Wasserstein examples as the radius of $L_{\infty}$ ball to generate $L_{\infty}$ PGD adversarial examples. For example, the maximum $L_{\infty}$ norm of the Wasserstein adversarial examples with 0.01 Wasserstein distance is 0.2. Then we will search for adversarial examples in the  $L_{\infty}$ ball with the radius equal to 0.2. In this way, we have a fair comparison between the Wasserstein PGD and the original PGD attack. 

\begin{figure}[ht]
\vspace{-0.2cm}
  \centering
  \setlength{\abovecaptionskip}{0.cm}
  \includegraphics[width=0.44\textwidth]{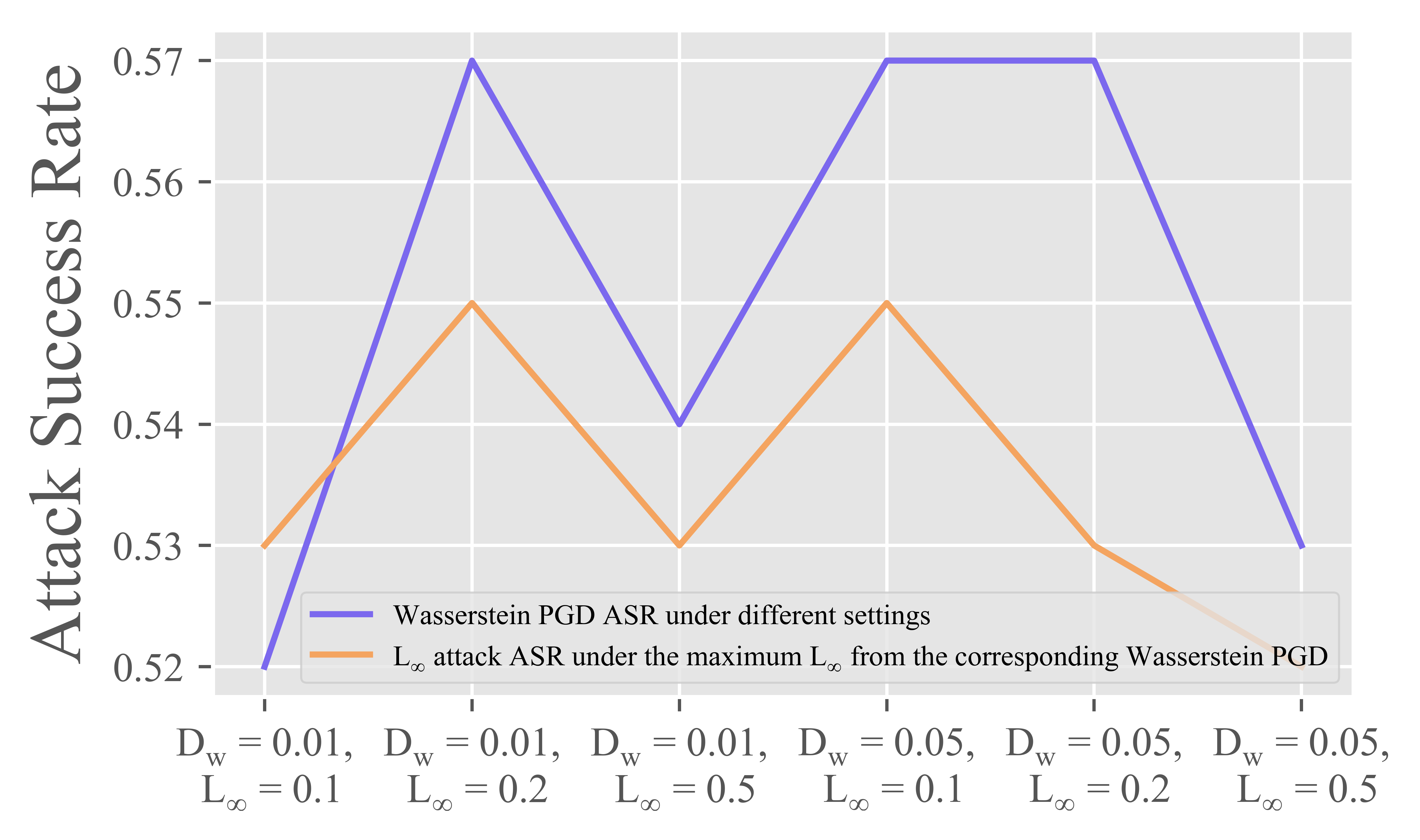}
  \caption{Comparison between Wasserstein PGD and $l_{\infty}$ PGD under the same attack scale.}
  \label{fig:compare2}
%   \vspace{-0.4cm}
\vspace{-2em}
\end{figure}

Figure \ref{fig:compare2} shows the comparison of ASR in the way we introduced above. The x-axis refers to different attack settings corresponding to different Wasserstein and $L_\infty$ bounds (note that we have two steps of projections, first to an $L_\infty$ norm ball and second to a Wasserstein ball, while original PGD only uses $L_\infty$ projection). The purple and orange lines correspond to the Wasserstein PGD and original PGD respectively. We can note that in most cases, Wasserstein PGD has a higher attack success rate than the original attack.

From these two aspects, we can conclude that for univariant time series data, Wasserstein PGD not only can generate more natural adversarial examples but also can achieve a higher ASR under the same attack scale.  

\begin{figure*}[ht]
 \centering
    \begin{subfigure}[b]{0.4\textwidth}
         \centering
         \includegraphics[width=\textwidth]{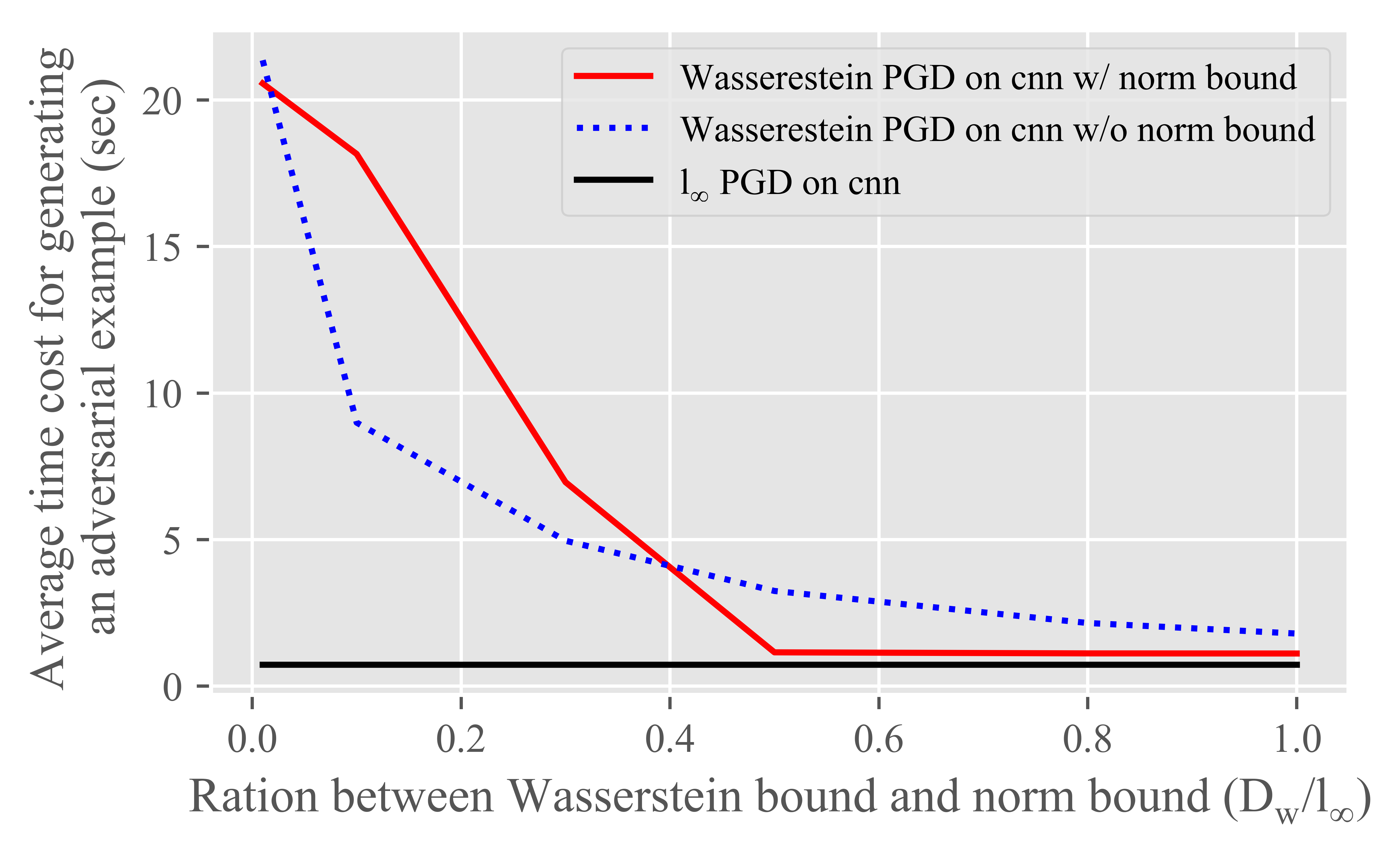}
         \caption{ECG200 CNN}
         \label{fig:ECG200_CNN}
     \end{subfigure}
     \begin{subfigure}[b]{0.4\textwidth}
         \centering
         \includegraphics[width=\textwidth]{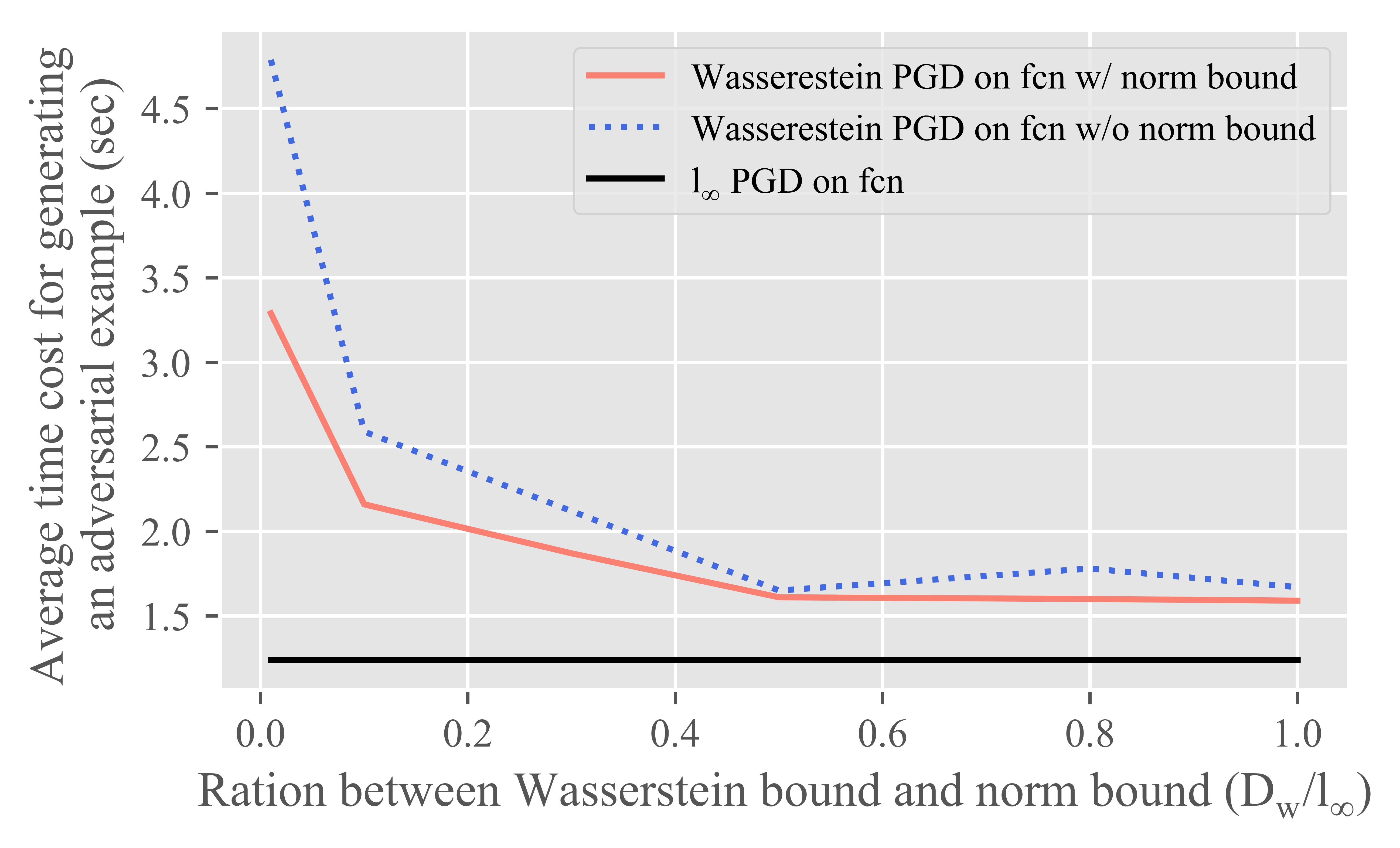}
         \caption{ECG200 FCN}
         \label{fig:ECG200_FCN}
     \end{subfigure}
     \begin{subfigure}[b]{0.4\textwidth}
         \centering
         \includegraphics[width=\textwidth]{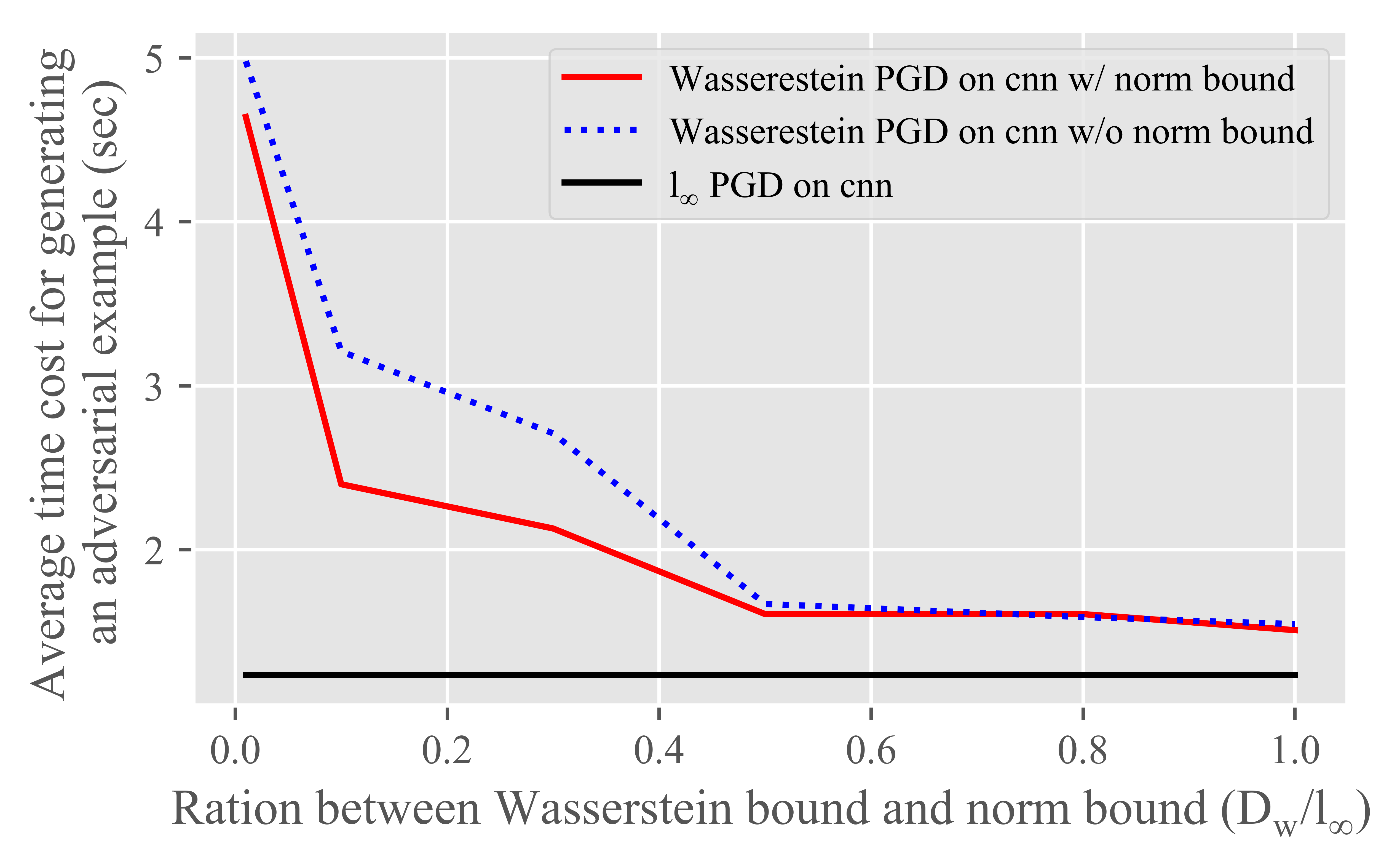}
         \caption{ECG5000 CNN}
         \label{fig:ECG5000_CNN}
     \end{subfigure}
     \begin{subfigure}[b]{0.4\textwidth}
         \centering
         \includegraphics[width=\textwidth]{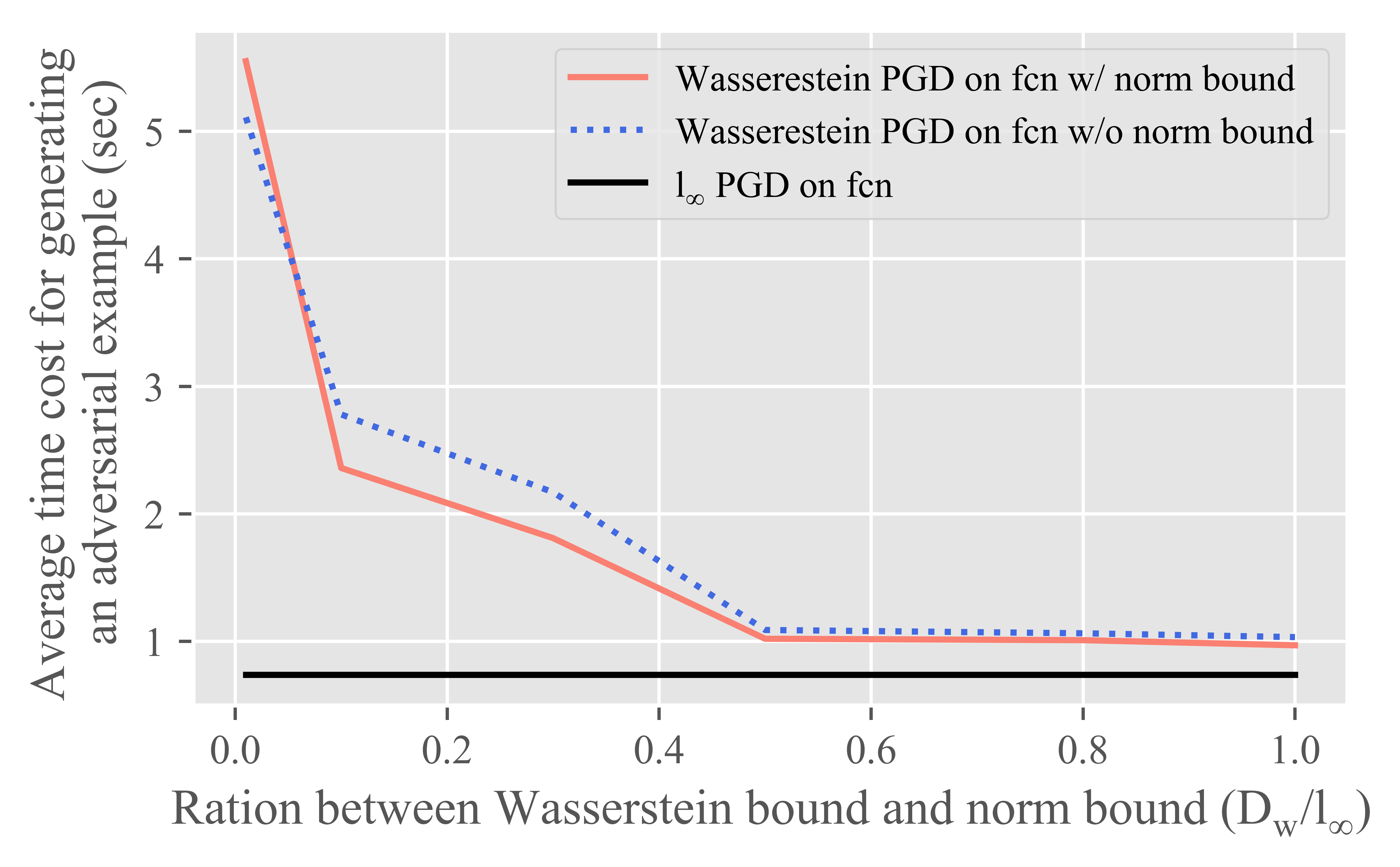}
         \caption{ECG5000 FCN}
         \label{fig:ECG5000_FCN}
     \end{subfigure}
\caption{The average time cost of generating an adversarial example with Wasserstien PGD (with and without norm bound clipping) and $L_{\infty}$ PGD attack.} 
\label{fig:time}
\vspace{-2em}
\end{figure*}

\subsubsection{Compare the time cost between Wasserstein PGD and $L_{\infty}$ PGD}
Besides comparing the utility of Wasserstein PGD with $L_{\infty}$ PGD, we also evaluate how Wasserstein PGD increases the time cost by comparing the average time cost of generating an adversarial example with $L_{\infty}$ PGD attack, as Wasserstein PGD involves projection into the Wasserstein ball via gradient which will result in more time cost. As the time cost of the Wasserstein projection is largely relative to the size of $L_{\infty}$ ball (the start point of gradient descent) and the size of the Wasserstein ball (the destination), we compare the average time cost according to the ratio between Wasserstein bound and $L_{\infty}$ bound.

As shown in Figure \ref{fig:time}, the red and blue curves are the Wasserstein PGD with and without the first stage of norm clipping, while the black line represents the baseline $L_{\infty}$ PGD attack, whose value is irrelevant to the ratio between Wasserstein and $L_{\infty}$ bound. We show the result on two datasets and two models: ECG200 and ECG5000, with CNN and FCN models. We can note that Wasserstein PGD will result in more time cost than $L_{\infty}$ PGD, especially when the ratio between Wasserstein bound and $L_{\infty}$ bound is large. However, when the ratio approaches 1, this increase in time cost is neglectable. By comparing the red and blue curves we can conclude that when the ratio between the Wasserstein bound and $L_{\infty}$ bound approaches 1, the first stage of norm clipping can effectively guide the search for Wasserstein adversarial examples. However, when the ratio increases( approaches 0), even with the bound of norm clipping, the search also tends to be a random search.

%%%%%%%%%%%%%%%%%%%%%%%%%%%%%%%%%%%%%%%%%%%%%%%%
%%%%%%%%%%%%%%%%%%%%%%%%%%%%%%%%%%%%%%%%%%%%%%%%
%%%%%%%%%%%%%%%%%%%%%%%%%%%%%%%%%%%%%%%%%%%%%%%%
\subsection{Countermeasure against Wasserstein PGD}
To better study the nature of Wasserstein adversarial examples, we also explored certified robustness approach as a potential defense mechanism for Wasserstein PGD. We consider certified robustness in contrast to other empirical defense methods as it is the most powerful and principled defense method to date. 

We adopt Wasserstein smoothing \cite{levine2020wasserstein} which is originally designed for image data not the univariant time series data.  The basic idea of Wasserstein smoothing is to define a reduced transport plan and map the Wasserstein distance on the input space to the $L_1$ norm on the transport plan. The base classifier is transformed into a smoothed classifier by adding Laplacian noise $Laplace(0, \sigma)^r$ to the reduced transport plan, where $r$ is corresponding to the input dimension. Because of the mapping, smoothing in the transport domain can be performed using existing $L_1$ robustness certification provided by \cite{PixelDP} and mapped back to the Wasserstein Space. The strict form of the certified condition can be stated as:
\begin{theorem}\label{theorem}
For any normalized probability distribution input $x \in \mathbb{R}^{n*m}$ with correct class $c$,  if:
\begin{equation}
    \bar{f_c}(x) \geq e^{2 \sqrt{2}\zeta / \sigma} (1-\bar{f_c}(x)) \label{condition}
\end{equation}
then for any perturbed $\tilde{x}$ that $d_W(x, \tilde{x}) \leq \zeta$, we have 
\begin{equation}
\bar{f_c}(\tilde{x}) \geq \underset{i \neq c}{max} \bar{{f_i}}(\tilde{x})
\end{equation}

The detailed proof and the design of the reduced transport plan can be referred to \cite{levine2020wasserstein}

\end{theorem}

We evaluate how Wasserstein Smoothing empirically works on the proposed Wasserstein PGD attack with two evaluation metrics \cite{wang2021certified}: \textbf{Certified accuracy (CertAcc)} which denotes the
fraction of the clean testing set on which the predictions are correct and also satisfy the certification criteria. 
 Formally, it is defined as:
$ \frac{\sum_{t = 1}^T certifiedCheck(X_t, L, \epsilon)\& corrClass(X_t, L,\epsilon)}{T},$
where {\small $certifiedCheck$} returns 1 if Theorem \ref{theorem} is satisfied and {\small $corrClass$} returns 1 if the classification output is correct. $T$ is the size of the test dataset. \textbf{Conventional accuracy (ConvAcc)} is defined as the fraction of testing set that is correctly classified, $\frac{\sum_{t = 1}^T corrClass(X_t, L,\epsilon)}{T}$, which is a standard metric to evaluate any deep learning systems.

\begin{figure}[ht]
\vspace{-0.1cm}
\setlength{\abovecaptionskip}{0.cm}
  \centering
  \includegraphics[width=0.45\textwidth]{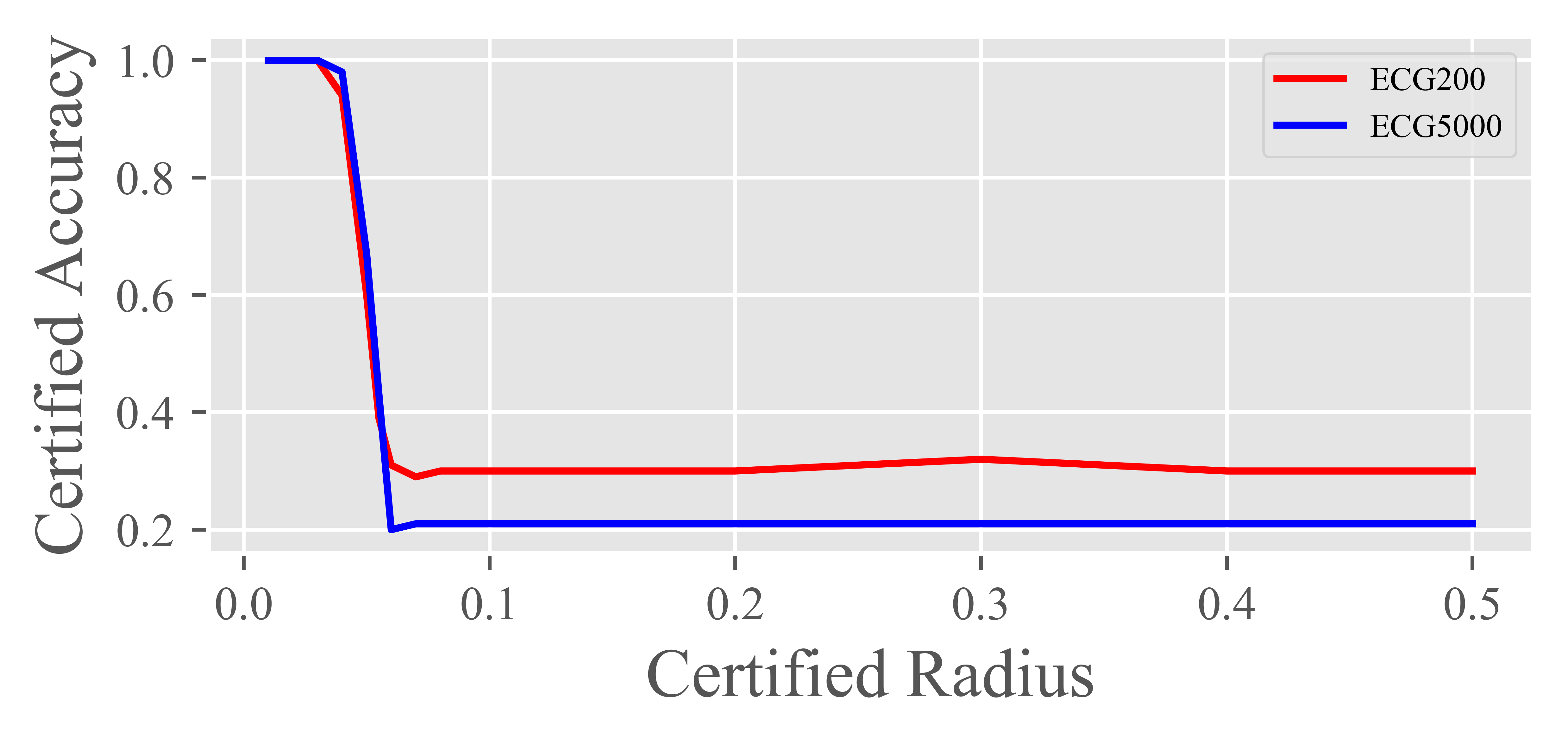}
  \caption{Comparison of Certified Accuracy under different Wasserstein radius with $\sigma = 0.01$}
  \label{fig:CertAcc}
\vspace{-0.2cm}
\end{figure}

Figure \ref{fig:CertAcc} illustrates how the certified accuracy changes under different Wasserstein certified radius $\zeta$ in Equation \ref{condition}. During the experiments, the Laplace parameter $\sigma$ which controls the noise to the transport plan is set to 0.01 and the soft prediction result is averaged over 100 times of sampling.  As Wasserstein radius increases, the certified accuracy decreases, which means less fraction of examples can satisfy the certification condition in Equation \ref{condition}. When  Wasserstein radius $\zeta$ is set over 0.1, the certified accuracy stops decreasing. We can note from this result that: first,  under a certain Laplacian distribution, the certified radius is limited. Second, to a very small certified radius, for examples less than 0.01, the certified accuracy can achieve 100\%.

\begin{figure}[ht]
\vspace{-0.1cm}
\setlength{\abovecaptionskip}{0.cm}
  \centering

  \includegraphics[width=0.45\textwidth]{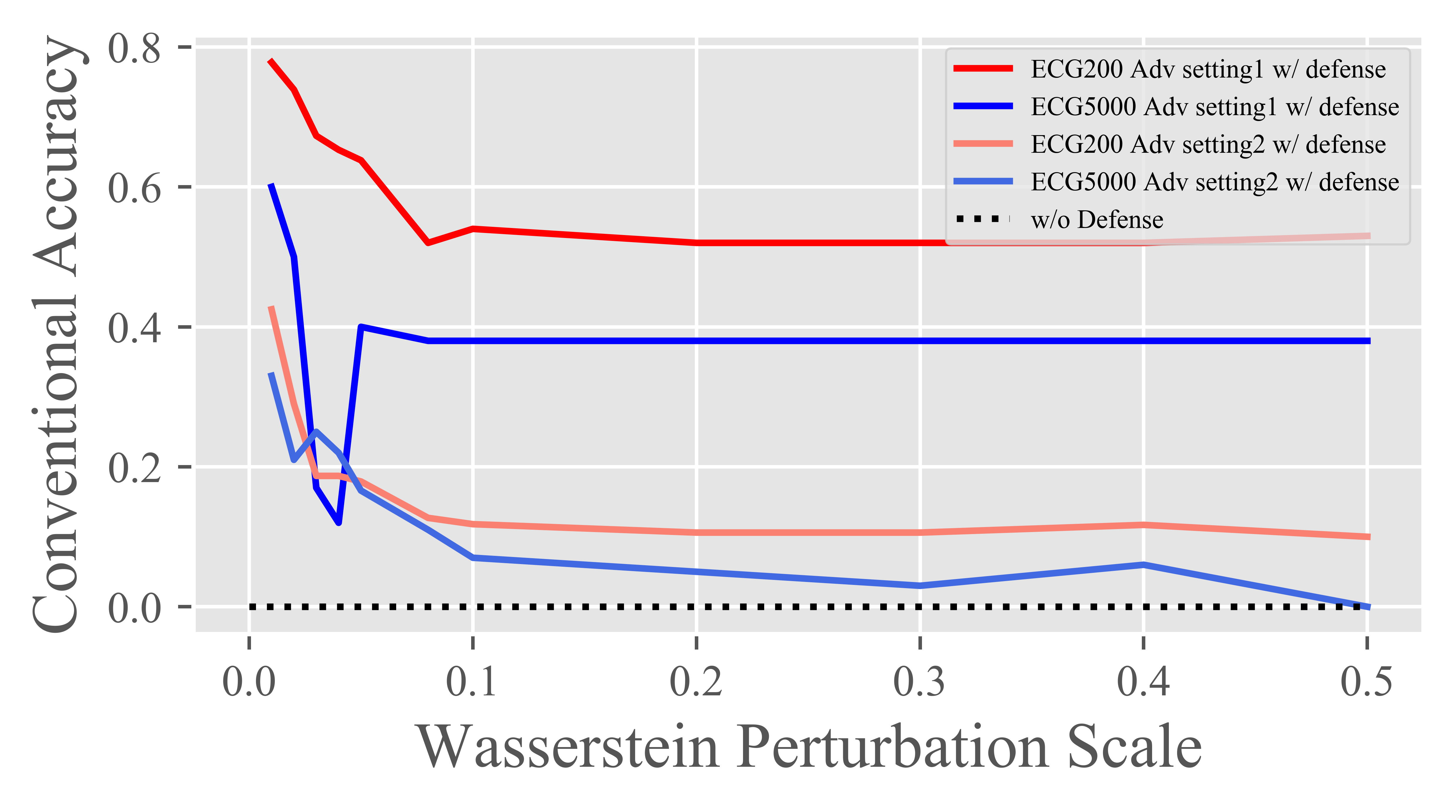}
  \caption{Comparison of Conventional Accuracy of successfully attacked adversarial examples. }
  \label{fig:ConvAcc}
\vspace{-0.2cm}

\end{figure}

\begin{figure}[ht]
 \centering
    \begin{subfigure}[b]{0.35\textwidth}
         \centering
         \includegraphics[width=\textwidth]{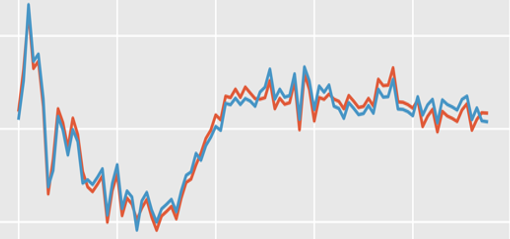}
     \end{subfigure}
     \begin{subfigure}[b]{0.35\textwidth}
         \centering
         \includegraphics[width=\textwidth]{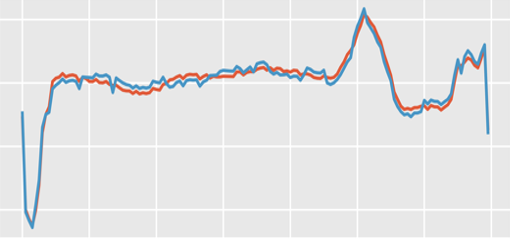}
     \end{subfigure}
\caption{The comparison between adversarial examples and clean examples at $d_\mathcal{W}$ scale around 0.06.} 
\label{fig:cert06}
\vspace{-1em}
\end{figure}

Figure \ref{fig:ConvAcc} demonstrates whether Wasserstein smoothing can empirically increase the conventional accuracy of adversarial examples generated by Wasserstein PGD. In the experiments, we test on successfully generated adversarial examples, so the baseline accuracy of the dataset is 0. Laplace parameter $\sigma$ is set to 0.1 and the soft prediction result is averaged over 100 times of sampling. The x-axis refers to the Wasserstein attack scales. The two red lines represent the adversarial examples generated from ECG200 under two different settings, $L_{\infty}$ norm set to 0.1 and 0.2, and the two blue lines represent the adversarial examples generated from ECG5000 where $L_{\infty}$ norm set to 0.1 and 0.2. We can note from the figures the following.  First, when the Wasserstein attack scales are small, Wasserstein smoothing can render some accuracy gain. The accuracy decreases with the increase of Wasserstein perturbation scales as expected.  When the Wasserstein distance increase over 0.06, which is also the greatest certified radius, the accuracy gain is limited and does not change anymore. We also randomly select two adversarial examples with Wasserstein perturbations around 0.06. As shown in Figure \ref{fig:cert06}, the adversarial ECG is fairly indistinguishable to human eyes. Yet Wasserstein Smoothing can not provide reasonable defense at this level. %also cannot facilitate the physician to tell the difference.  
Second, with ECG200 and ECG5000 under setting2 ( $L_{\infty}$ norm set to 0.2), the overall accuracy gain is very small for all perturbation scales. 

We can conclude from the results that the existing Wasserstein smoothing has limited success in both certified ratio and conventional accuracy gain. This suggests that there is still space for developing stronger certified robustness method to Wasserstein PGD tailored to time series data instead of using the general transport plan based smoothing designed for image data.

%% file: conclusion.tex
\section{Conclusion}

We proposed the Wasserstein PGD attack for univariant time series data, which is the first effort to study adversarial example attacks on time series in the Wasserstein space. Compared with the original PGD attack in the Euclidean space, Wasserstein PGD can generate more imperceptible and natural adversarial examples. Extensive results showed that Wasserstein PGD can achieve higher ASR with smaller perturbations. We also explored certified robustness via Wasserstein smoothing as a potential defense method. The results showed that Wasserstein smoothing has limited certified range and accuracy gain. In the future, it would be interesting to study Wasserstein adversarial examples on multivariant time series under Gaussian distributions. Stronger certified robustness methods can also be developed by designing reduced transport plan that is more suitable for time series data. 